\pdfoutput=1
\documentclass[11pt]{article}

\usepackage{acl}

\usepackage{times}
\usepackage{latexsym}
\usepackage[T1]{fontenc}
\usepackage[utf8]{inputenc}
\usepackage{microtype}
\usepackage{inconsolata}

\usepackage{graphicx}
\usepackage{booktabs}
\usepackage{multirow}
\usepackage{array}
\usepackage{colortbl}
\usepackage{makecell}
\usepackage{xcolor}
\usepackage{amsmath,bm}
\usepackage{amssymb}
\usepackage{amsfonts}
\usepackage[hang,flushmargin]{footmisc}
\usepackage[most]{tcolorbox}
\usepackage{subcaption}
\usepackage{algorithm}
\usepackage{algorithmic}
\usepackage{wrapfig}
\usepackage{listings}

\title{Unlocking Fine-Grained and Within-Utterance Speaking Style Control\\
in Text-to-Speech Models}

\author{Jaehoon Kang, Yejin Lee, Yoonji Park, Kyuhong Shim \\
Sungkyunkwan University, Republic of Korea \\
\texttt{morateng@skku.edu, khshim@skku.edu}
}

\begin{document}
\maketitle

\begin{abstract}
Modern text-to-speech (TTS) models enable natural-language-driven style control but often provide limited fine-grained control and apply a single global style per utterance.
We propose training-free inference-time methods for continuous inter-utterance interpolation and time-varying intra-utterance transitions.
For inter-utterance control, we interpolate direction vectors between contrastive style prompts in the embedding space.
For intra-utterance control, we identify an early-token attention bias that causes the initial audio realization to dominate subsequent generation and mitigate it using KV-cache swapping and sliding-window attention masking.
Experiments demonstrate 99--100\% gender conversion success, up to 36 Hz pitch variation, and up to 1.6 syllables-per-second speed change.
Specifically, our intra-utterance transition maintains speaker similarity of 0.81--0.91 and achieves perceptual smoothness scores of 3.48--4.48.
\end{abstract}

\section{Introduction}\label{sec:intro}

Recent advances in text-to-speech (TTS) synthesis have introduced prompt-conditioned models that accept natural-language descriptions of speaking style~\cite{guo2023prompttts,yang2024instructtts,lacombe-etal-2024-parler-tts}.
This approach offers a flexible interface for convenient and expressive voice generation; instead of preparing reference audio or selecting from a fixed set of style presets, users can describe intent directly in text (e.g., ``calm male voice'', ``fast and high-pitched speech'').

\begin{figure}[t]
    \centering
    \includegraphics[width=1.0\linewidth]{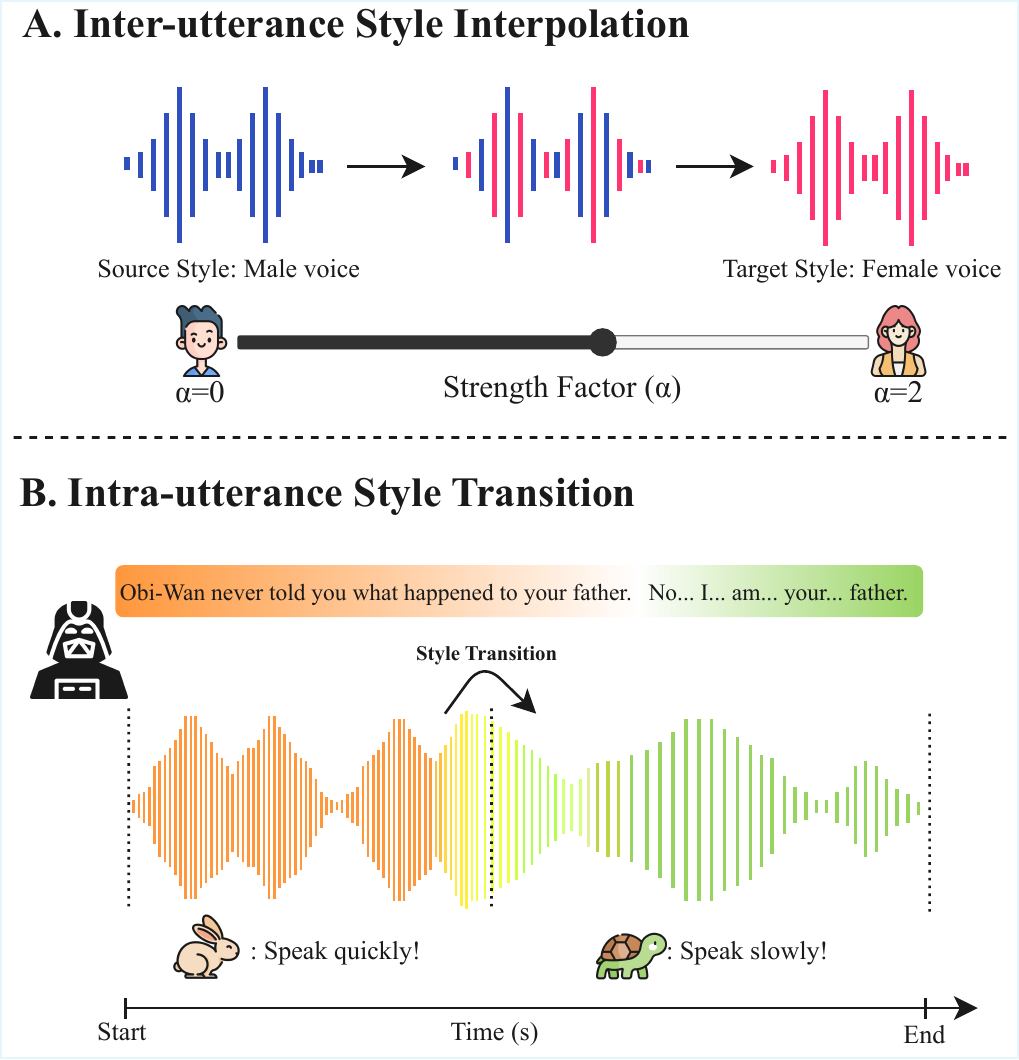}
    \caption{Overview of our training-free style control methods. (A) \textbf{Inter-utterance style interpolation} provides continuous control between source and target styles by adjusting the strength factor $\alpha$. 
    (B) \textbf{Intra-utterance style transition} varies speaking style within a single generated utterance.}
    \label{fig:overview}
    % \vspace{-0.2cm}
\end{figure}

Despite this progress, achieving fine-grained and predictable control from natural language style prompts remains challenging~\cite{huang2025instructttseval}.
Many speech attributes, such as pitch and speed, vary continuously.
However, current style-conditioned TTS models mostly accept coarse, discrete categorical terms (e.g., ``fast'', ``slightly fast'', ``very fast'', etc.) and small prompt edits do not reliably produce smooth, monotonic, and predictable changes in acoustic attributes~\cite{korotkova2024word,ji2025controlspeech}.
This is partially because current models assume a fixed global style condition~\cite{leng2024prompttts2,yang2024instructtts,ji2024textrolspeech}, generating speech that is stylistically consistent but difficult to steer at a finer resolution.

These limitations are critical because many real-world applications require both continuous and time-varying control.
Audiobook narration benefits from gradual changes in speed or pitch to convey tension and emphasis~\cite{guo2024audiobook}.
Conversational agents may need to shift tone within a single response to match discourse structure~\cite{liu2024emotion}.
Motivated by such use cases, we focus on two complementary controllability goals that are not well supported by current style-controllable TTS models: (1) \textit{inter-utterance style interpolation}: continuous control between contrastive style attributes (e.g., selecting an intermediate speaking rate between fast and slow) and (2) \textit{intra-utterance style transition}: style transition within a single utterance (e.g., starting fast and gradually slowing down).

In this paper, we propose training-free methods that achieve both continuous inter-utterance controllability and intra-utterance style transition for natural-language-based TTS, as illustrated in Figure~\ref{fig:overview}.
We discover that existing models implicitly contain such controllability that can be \textit{unlocked} via inference-time interventions.
For inter-utterance control, we show that a simple representation-level approach is sufficient.
Interpolating between style embeddings induced by contrastive prompts allows for finer control than discrete prompt edits.

In contrast, intra-utterance style transition reveals a qualitatively different obstacle.
We identify a previously unreported phenomenon, termed \textbf{style self-referencing}: the first few seconds of generated speech disproportionately govern subsequent generation and can substantially reduce the effect of the natural-language style prompt. 
Consequently, na\"{\i}ve inference-time solutions, such as interpolating style embeddings or swapping prompts during generation, fail to produce reliable intra-utterance transitions.

Based on this analysis, we introduce novel inference-time mechanisms that directly counteract this self-referencing effect.
Specifically, we employ (1) key-value (KV) cache swap and (2) sliding-window self-attention masking, both aiming to reduce the dominance of early-generated tokens.
Together, these unlock smooth intra-utterance style transitions without fine-tuning.

\vspace{0.1cm}
Our contributions are as follows:
\vspace{-0.1cm}
\begin{itemize}
    \item We introduce two complementary forms of style controllability in natural language-conditioned TTS: inter-utterance style control and intra-utterance style transition.
    \vspace{-0.1cm}
    \item We identify and characterize self-referencing in autoregressive TTS generation, where early-generated speech dominates later segments and overrides natural language guidance.
    \vspace{-0.1cm}
    \item Our proposed methods are entirely training-free, achieving fine-grained style control through inference-time techniques.
    \vspace{-0.1cm}
    \item We provide comprehensive evaluation demonstrating effective continuous control over pitch, speed, and gender attributes while maintaining speech quality.
\end{itemize}
\section{Related Work}\label{sec:related}

\subsection{Style Controllable Text-to-Speech}
The pursuit of controllable TTS has primarily evolved through two paradigms: reference-based conditioning and, more recently, natural language-based prompting~\cite{xie2025towards}.

\paragraph{Reference-based approaches.}
Early TTS models conditioned synthesis on reference audio to capture style attributes. 
Global Style Tokens (GST)~\cite{wang2018style} and VAE-based methods~\cite{zhang2019learning, hsu2018hierarchical} model prosodic variations as continuous latent variables, enabling style interpolation in the latent space.
More recently, large-scale autoregressive models such as VALL-E~\cite{wang2023neural,chen2024valle2}, NaturalSpeech 3~\cite{ju2024naturalspeech}, CosyVoice~\cite{du2024cosyvoice,du2025cosyvoice}, and Spark-TTS~\cite{wang2025spark} achieve high-fidelity zero-shot synthesis by conditioning on reference speech.
Despite their fidelity, these approaches require reference audio, which prevents users from synthesizing specific styles in the absence of matching acoustic samples.

\begin{figure*}[t!]
    \centering
    \begin{subfigure}[b]{0.58\linewidth}
        \centering
        \includegraphics[width=\linewidth]{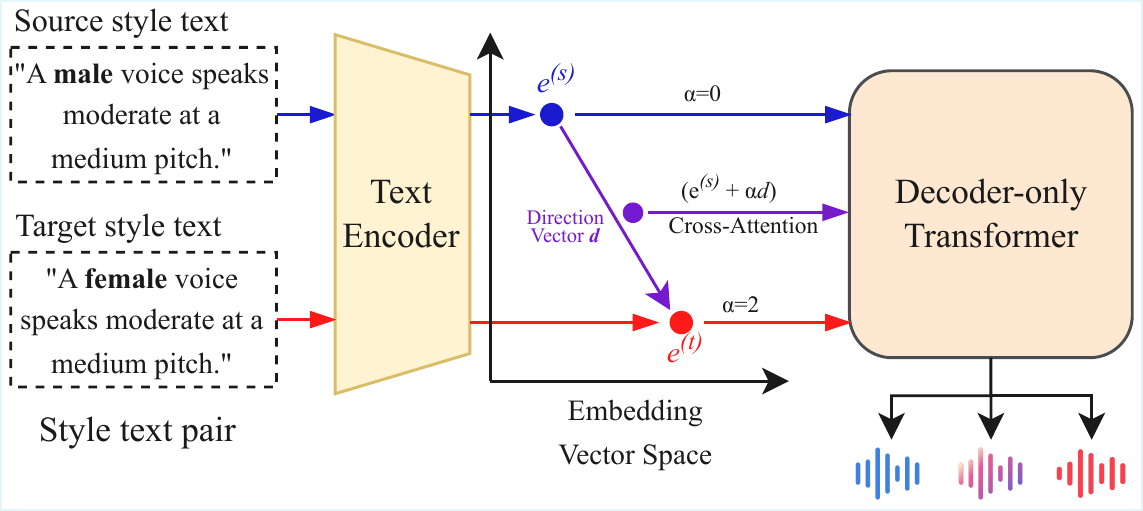}
        \caption{Overview of the inter-utterance style interpolation.}
        \label{fig:inter_method}
    \end{subfigure}
    \hfill
    \begin{subfigure}[b]{0.38\linewidth}
        \centering
        \includegraphics[width=\linewidth]{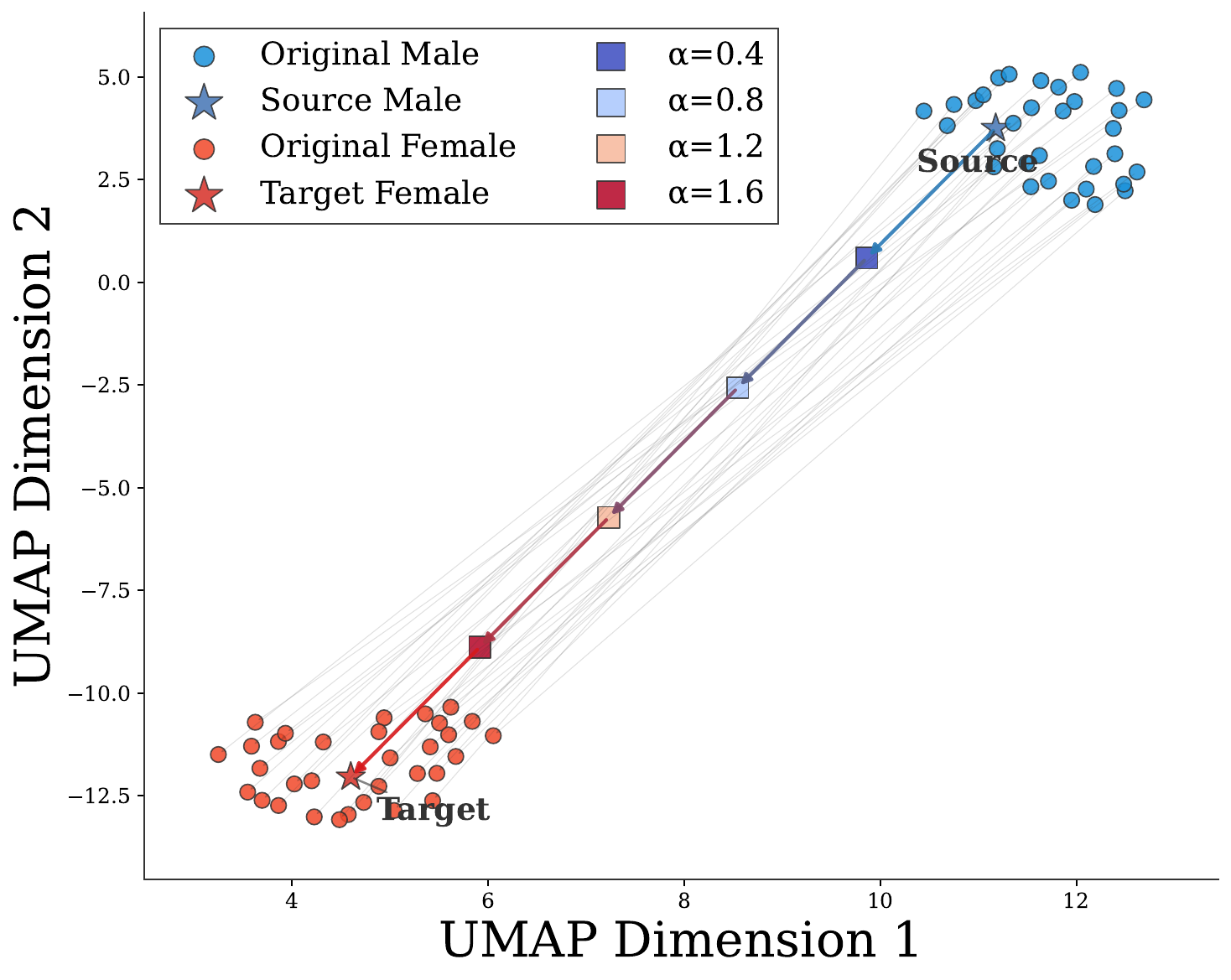}
        \caption{U-MAP visualization.}
        \label{fig:umap}
    \end{subfigure}
    \vspace{0.2cm}
    \caption{\textbf{Inter-utterance Style Interpolation.} 
    (a) Given a source-target style prompt pair, we extract their embeddings 
    $\bm{E}^{(s)}$ and $\bm{E}^{(t)}$ from the text encoder and compute the direction vector 
    $\bm{d}_i = \frac{1}{2}(\bm{e}^{(t)}_i - \bm{e}^{(s)}_i)$ for attribute-token positions $i \in \mathcal{A}$. 
    By varying the interpolation strength $\alpha$, we generate new style embeddings that can continuously control the speaking style attributes. 
    (b) UMAP visualization of the embedding space shows that interpolated embeddings form a smooth trajectory between source (male) and target (female) style representations.}
    \label{fig:temporary}
\end{figure*}

\paragraph{Natural language-based approaches.}
To overcome the reliance on reference audio, recent research has shifted towards controlling TTS via natural language descriptions~\cite{liu2023promptstyle}.
PromptTTS~\cite{guo2023prompttts} pioneers this by learning a mapping between textual descriptions and acoustic style latents, enabling attribute control (e.g., gender, pitch, speed) through prompts. 
This line of work has been rapidly extended by models such as InstructTTS~\cite{yang2024instructtts}, which employs cross-modal metric learning to follow free-form instructions, and Parler-TTS~\cite{lyth2024natural}, which leverages large-scale synthetic annotations to achieve high-fidelity description-guided synthesis. 
However, these models, as well as more recent systems such as FlexiVoice~\cite{chen2026flexivoice} and InstructAudio~\cite{qiang2026instructaudio} that improve instruction adherence through post-training or unified speech-audio generation, generally treat style prompts as global, static characteristics applied uniformly across the entire utterance.
They lack mechanisms to interpret time-varying prompts (e.g., ``start calmly and become excited'') or to continuously modulate style intensity within an utterance.

\subsection{Fine-Grained Speaking Style Control}

Significant efforts have progressed from predicting prosodic features via variance adaptors in non-autoregressive models like FastSpeech-2~\cite{ren2021fastspeech} and FastPitch~\cite{lancucki2021fastpitch} to more explicit manipulation of intra-utterance dynamics~\cite{chen2022fine, wang2023fine}. 
Recent advancements such as Lina-Style~\cite{lemerle2025lina} and WeSCon~\cite{wang2025wordlevel} achieve word-level emotional control through synthetic data interleaving and self-training with multi-round inference, respectively.
Furthermore, models like ELaTE~\cite{kanda2024making} and EmoCtrl-TTS~\cite{wu2024laugh} extend this control to non-verbal vocalizations and continuous arousal-valence trajectories, allowing for a more dynamic emotional flow than traditional text-driven generations.

The integration of LLMs allows intuitive, zero-shot emotion control through natural-language prompts, as explored in PUE~\cite{gao2025prompt}. 
However, these prior efforts differ fundamentally from ours in both backbone and scope.
EmoSteer-TTS~\cite{xie2025emosteer} constructs steering directions from neutral--emotional speech pairs and modifies the residual streams of flow-matching TTS models for global emotion control.
In contrast, we use \emph{contrastive text prompts} and intervene in autoregressive decoders through KV-cache swapping and sliding-window masking to enable \textit{intra-utterance} style transitions.
Here, the style self-referencing phenomenon emerges as a distinct challenge in autoregressive generation.
Recent work further explores intermediate representations~\cite{wang2025controllable} and instruction-guided editing datasets~\cite{chen2026isse}, but operates at the data or representation level rather than directly intervening in the decoder at inference time.
Consequently, achieving natural intra-utterance style transitions remains an open challenge.

\section{Inter-Utterance Style Interpolation}~\label{sec:inter_utterance}
In this section, we present our method to achieve continuous inter-utterance style control by manipulating direction vectors between contrastive style prompts in the style embedding space.

\begin{figure*}[t!]
    \centering
    \includegraphics[width=\textwidth]{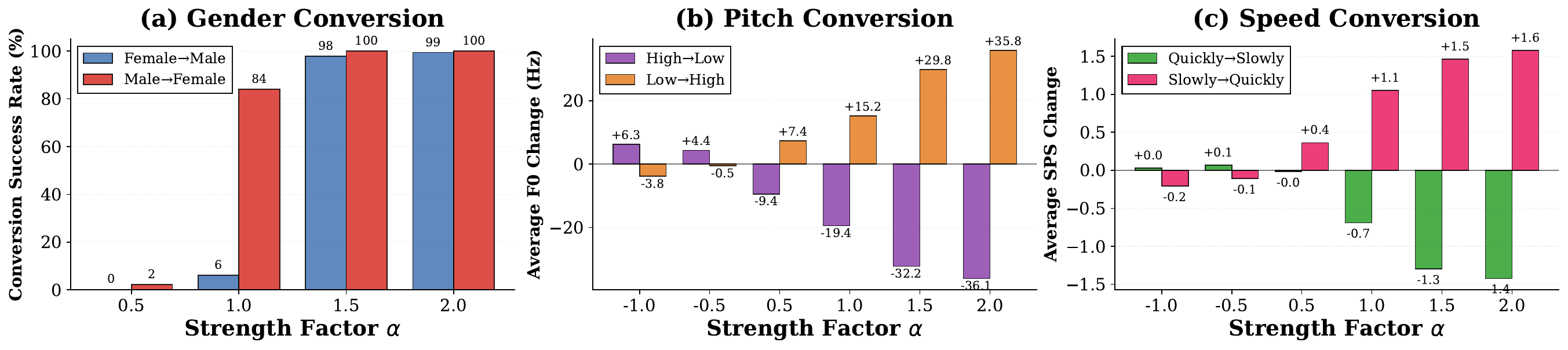}
    \caption{Inter-utterance style interpolation results. (a) Gender conversion success rate, (b) average F0 change for pitch control, (c) average speaking rate change for speed control. 
    All attributes show monotonic changes with $\alpha$ in the positive range, demonstrating continuous controllability.}
    \label{fig:inter_utterance}
\end{figure*}

\begin{table*}[t]
\centering
\vspace{0.2cm}
\setlength{\tabcolsep}{5pt}
\small
\begin{tabular*}{\textwidth}{@{\extracolsep{\fill}}llcccccccc@{}}
\toprule
\multirow{2}{*}{\textbf{Attr.}} & \multirow{2}{*}{\textbf{Direction}} & \multicolumn{3}{c}{\textbf{Objective}} & \multicolumn{3}{c}{\textbf{Style Change ($\alpha$)}} & \multirow{2}{*}{\textbf{MOS}} \\
\cmidrule(lr){3-5} \cmidrule(lr){6-8}
 & & \textbf{Succ.} & $\boldsymbol{\Delta}$ \textbf{Metric} & \textbf{SIM} & $0.5$ & $1.0$ & $2.0$ & \\
\midrule
\multirow{2}{*}{Gender} & F $\rightarrow$ M & 99.0\% & -- & -- & 0.09 & 0.12 & 1.74 & 4.25 \\
                        & M $\rightarrow$ F & 100.0\% & -- & -- & 0.18 & 0.62 & 2.00 & 4.40 \\
\midrule
\multirow{2}{*}{Pitch}  & H $\rightarrow$ L & 96.3\% & $-$36.1 Hz & 0.76 & 0.95 & 1.09 & 1.71 & 4.26 \\
                        & L $\rightarrow$ H & 93.0\% & $+$35.8 Hz & 0.78 & 0.68 & 1.18 & 1.62 & 4.29 \\
\midrule
\multirow{2}{*}{Speed}  & Q $\rightarrow$ S & 94.2\% & $-$1.4 SPS & 0.84 & 0.26 & 0.75 & 1.62 & 3.99 \\
                        & S $\rightarrow$ Q & 95.7\% & $+$1.6 SPS & 0.84 & 0.43 & 1.45 & 1.77 & 4.32 \\
\bottomrule
\end{tabular*}
\caption{Inter-utterance style interpolation results. \textbf{Objective}: success rate, $\Delta$ metric (Hz for pitch, SPS for speed) at $\alpha = 2.0$, and speaker similarity (SIM). \textbf{Style Change}: subjective ratings ($-2$ to $+2$) at different interpolation strengths $\alpha$, with MOS for naturalness ($\alpha = 2.0$).}
\label{tab:inter_utterance}
\end{table*}

\subsection{Analysis: Style Vectors in Embedding Space}

To investigate the characteristics of natural language style embeddings, we analyze how contrastive style attributes are represented in the embedding space.
Specifically, we prepare multiple pairs of style prompts that differ only in a single target attribute. For example, for gender control, we use pairs like ``A \textbf{male} voice with clean audio'' and ``A \textbf{female} voice with clean audio'', where only the gender token varies while other attributes in the prompt remain identical. 
We encode these prompts using the prompt text encoder and extract only the embeddings corresponding to the target style attribute tokens (e.g., ``male'' and ``female'').
To visualize their distribution, we project these attribute embeddings into a lower-dimensional space using UMAP~\cite{mcinnes2018umap}.

As shown in Figure~\ref{fig:temporary}b, we observe that embeddings of the same attribute form tight clusters: all ``male'' tokens cluster together, and all ``female'' tokens form a separate cluster. Importantly, these contrastive attribute clusters (e.g., ``male'' vs. ``female'') are well-separated in the embedding space~\cite{kang2026encoderfree}. This clear separation suggests that linear interpolation between contrastive attribute vectors can produce continuous style transitions~\cite{zou2023representation, panickssery2023steering, park2024linear}.

\subsection{Method}

We implement a simple embedding vector interpolation method for continuous style control.
Let $\bm{E}^{(s)} = \{\bm{e}^{(s)}_1, \bm{e}^{(s)}_2, ..., \bm{e}^{(s)}_l\}$ and $\bm{E}^{(t)} = \{\bm{e}^{(t)}_1, \bm{e}^{(t)}_2, ..., \bm{e}^{(t)}_l\}$ denote the text encoder outputs for the source and target style prompts, respectively.
Since the two prompts differ only in the attribute tokens, we compute the direction vector for the set of attribute token positions, $\mathcal{A}$.
Then, the direction vector is computed as follows:
\begin{equation}
    \bm{d}_i = \frac{1}{2}\Big(\bm{e}^{(t)}_i - \bm{e}^{(s)}_i\Big), \quad i \in \mathcal{A}
\end{equation}
We apply the direction vector only to the positions corresponding to the attribute tokens.
The interpolated embedding is computed as:
\begin{equation}
    \bm{e}'_i = \begin{cases}
        \bm{e}^{(s)}_i + \alpha \cdot \bm{d}_i & \text{if } i \in \mathcal{A} \\
        \bm{e}^{(s)}_i & \text{otherwise}
    \end{cases}
    \label{eq:interpolation}
\end{equation}
where $\alpha \in \mathbb{R}$ is the interpolation strength.
When $\alpha = 0$, the output reproduces the source style; when $\alpha = 2$, it produces the target style; intermediate values yield smoothly interpolated styles.
Values outside $[0, 2]$ correspond to extrapolation beyond the original style range.

Figure~\ref{fig:inter_method} illustrates our method.
Given a source-target style text pair, we extract their embeddings from the text encoder and compute the direction vector between the contrastive tokens. 
By adding the scaled direction vector $\alpha \cdot\bm{d}$ to the source embedding, we obtain the interpolated style representation $\bm{E}'$.
This representation is then fed to the decoder through cross-attention to generate speech with the desired style characteristics.
We empirically verify that applying interpolation to all tokens yields similar performance to our attribute-only approach (see Appendix~\ref{appendix:full_vs_attribute}).

\subsection{Experimental Results}

\paragraph{Setup.}
We evaluate our inter-utterance style interpolation on three attributes: gender, pitch, and speed.
We take 400 sentences from the LibriTTS-R~\cite{koizumi2023libritts} test-clean split and generate speech using the Parler-TTS-mini model~\cite{lacombe-etal-2024-parler-tts, lyth2024natural}.
Further details are provided in Appendix~\ref{app:exp_details}.

\paragraph{Objective Evaluation.}
Figure~\ref{fig:inter_utterance} demonstrates that our method achieves effective style control across all tested attributes.
For gender conversion, the success rate increases with $\alpha$, achieving near-complete conversion at $\alpha\geq 1.5$ (98\% for Female$\rightarrow$Male, 100\% for Male$\rightarrow$Female).
For pitch and speed, which are continuous attributes, our method enables smooth and gradual control.
Pitch (F0) changes linearly with $\alpha$, achieving up to 36.1 Hz decrease (High$\rightarrow$Low) and 35.8 Hz increase (Low$\rightarrow$High) at $\alpha=2.0$, with intermediate values producing perceptibly distinct pitch levels.
Similarly, speaking rate varies proportionally with $\alpha$; Quick$\rightarrow$Slow reduces syllable-per-second (SPS)~\cite{wang2025spark} by 0.7 at $\alpha=1.0$ and 1.4 at $\alpha=2.0$, while Slow$\rightarrow$Quick increases SPS by 1.1 and 1.6, respectively, demonstrating fine-grained continuous control.
Table~\ref{tab:inter_utterance} summarizes the results at $\alpha = 2.0$.
Appendix~\ref{appendix:extrapolation} analyzes wider extrapolation and its eventual saturation and failure modes.

\begin{figure}[t!]
    \centering
    \includegraphics[width=\linewidth]{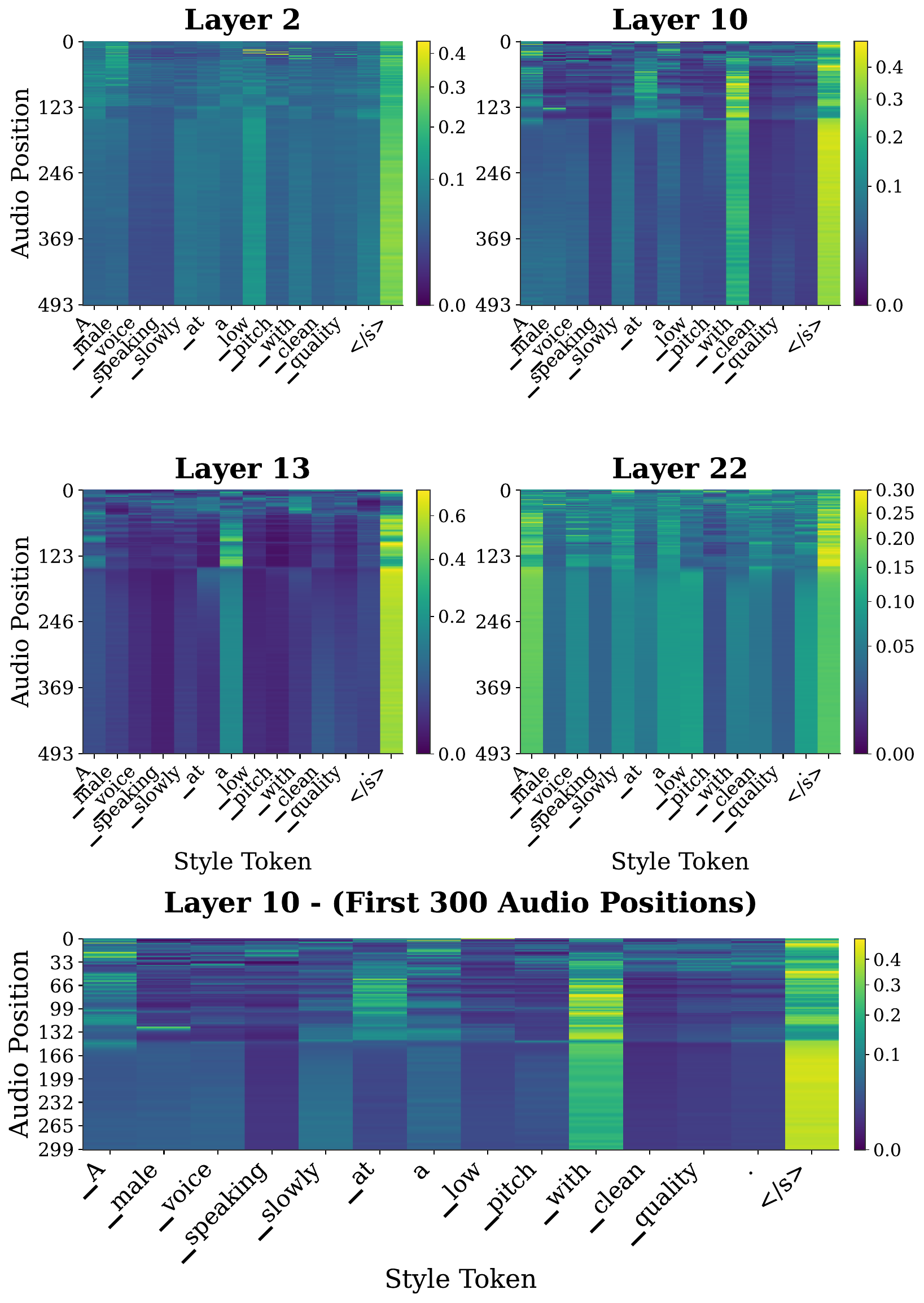}
    \caption{Cross-attention weights between style text tokens (rows) and generated audio tokens (columns) across multiple decoder layers. Attention weights are actively updated during early generation, then remain nearly constant throughout subsequent generation.}
    \label{fig:attn_map}
\end{figure}

\begin{figure*}[t!]
    \centering
    \includegraphics[width=\linewidth]{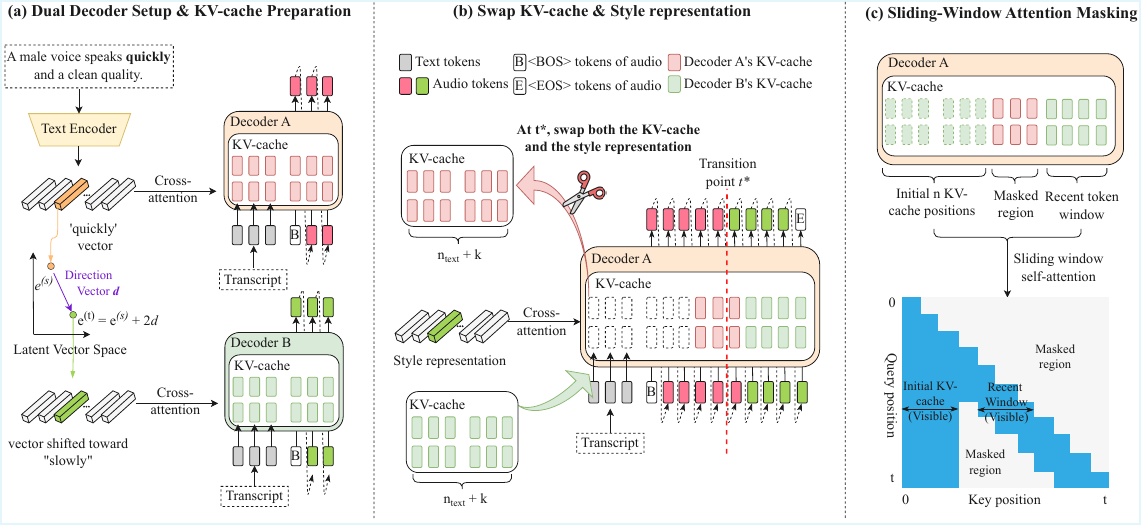}
    \caption{\textbf{Intra-utterance style transition.} To convert speaking style mid-generation, we combine multiple techniques to avoid style self-referencing: (a) prepare an initial KV-cache derived from the target style prompt, and (b) swap the original KV-cache and style embeddings, together with (c) sliding-window attention masking.}
    \label{fig:intra_method}
\end{figure*}

\paragraph{Subjective Evaluation.} 
In Table~\ref{tab:inter_utterance}, we show human evaluation results in which participants rated both style change ($-2$ to $+2$) and naturalness (MOS, 1--5) (see Appendix~\ref{app:inter_details} for details).
For continuous attributes (pitch and speed), participants consistently rated intermediate interpolation strengths ($\alpha = 0.5, 1.0$) lower than full conversion ($\alpha = 2.0$), confirming gradual style transitions.
At $\alpha = 2.0$, all conversion directions achieved strong perceived style change (scores 1.62--2.00) while maintaining high naturalness (MOS above 3.99).

\section{Intra-utterance Style Transition}~\label{sec:intra_utterance}
In this section, we address the novel challenge of varying style within a single utterance during autoregressive generation. 
We first describe our initial attempts and analyze why they fail, then present our solution.

\subsection{Intuition: Why Na\"{\i}ve Approaches Fail}

Building upon our inter-utterance style interpolation (Section~\ref{sec:inter_utterance}), we initially attempted to achieve intra-utterance style variation during autoregressive generation in a similar manner. 
Specifically, we expected that switching the style embedding from $\bm{E}^{(s)}$ to $\bm{E}^{'}$ at a transition point $t^*$ would cause the autoregressive decoder to generate all subsequent tokens ($t > t^*$) with the target style.

However, this na\"{\i}ve approach fails to produce desired style transitions.
In fact, the generated speech \textit{continues to exhibit the initial style characteristics} even after modifying the style embeddings.
This unexpected behavior, where the same manipulation that works across utterances fails within a single utterance, motivated us to investigate the decoder's attention patterns.

\paragraph{Early-Token Bias in Cross-Attention.}~\label{sec:early_bias}
To understand why modifying style representations mid-generation fails, we analyze the cross-attention patterns between generated audio tokens and style text tokens. 
Figure~\ref{fig:attn_map} visualizes the cross-attention weights in the autoregressive decoder across multiple layers, where rows correspond to style text tokens and columns correspond to audio token positions during generation.

In the early generation phase, the decoder actively attends to style tokens with dynamically changing attention patterns. 
This indicates that the model incorporates style information from the text prompt to establish the acoustic characteristics of the speech.
In contrast, in the following generation phase, the attention weights across all style tokens become fixed and show minimal variation. 
Rather than continuing to query the style representation, the decoder maintains a consistent attention distribution focusing on relatively uninformative (e.g., ``with'', ``<EOS>'') tokens.

This observation suggests that style-controllable autoregressive TTS models follow a \textit{set-and-maintain} strategy: the model establishes style characteristics during initial generation and subsequently maintains consistency by relying on these early acoustic representations through autoregressive (i.e., self-attention) conditioning~\cite{gu2024attention, barbero2025llms, xiao2024sink}.
We term this phenomenon \textbf{style self-referencing}.
Once the style is encoded in the early audio tokens, the decoder becomes resistant to new style information provided through cross-attention.

This early-token attention bias explains why our initial approach fails: even when we replace the style representation at $t^*$, the decoder still relies on the source style cached in early tokens. 
Therefore, achieving intra-utterance style variation requires not only modifying the cross-attention style input but also \textit{addressing the persistent influence of early-generated tokens} on autoregressive generation.

\subsection{Method}
Motivated by this analysis, we combine KV-cache swap and sliding-window attention masking to enable intra-utterance style transitions.
Figure~\ref{fig:intra_method} illustrates the overall process.

\paragraph{Dual Decoder Setup.}
To enable style transitions during generation, we prepare two decoder instances.
\textit{Decoder-A} generates speech in source style, conditioned on the original style representation $\bm{E}^{(s)}$ until the desired transition point $t^{*}$.
In parallel, \textit{Decoder-B} independently samples the first $k$ audio tokens to construct a target-style KV-cache, conditioned on the modified representation $\bm{E}'$ obtained through vector interpolation (Section~\ref{sec:inter_utterance}).
We report the measured inference overhead of this dual-decoder setup in the experimental results below.

\begin{table*}[t]
\centering
\small
\renewcommand{\arraystretch}{1.1}
\begin{tabular*}{\textwidth}{@{\extracolsep{\fill}}llccccc}
\toprule
\textbf{Attribute} & \textbf{Direction} & \textbf{Window Size} & \textbf{$\Delta$ Metric} & \textbf{SIM} & \textbf{Trans. (\%)} & \textbf{TMOS} \\
\midrule
\multirow{8}{*}{Pitch} & \multirow{4}{*}{High $\rightarrow$ Low} & 256 & \textbf{$-$12.4 Hz} & 0.81 & \textbf{96.2} & 4.20 \\
 & & 384 & $-$8.97 Hz & 0.85 & 73.1 & 3.79 \\
 & & 512 & $-$10.9 Hz & 0.87 & 80.0 & 4.20 \\
 & & Full & $-$11.5 Hz & \textbf{0.90} & 55.6 & 4.00 \\
\cmidrule{2-7}
 & \multirow{4}{*}{Low $\rightarrow$ High} & 256 & \textbf{$+$27.4 Hz} & 0.84 & \textbf{96.2} & 3.48 \\
 & & 384 & $+$19.6 Hz & 0.87 & 73.1 & 3.79 \\
 & & 512 & $+$16.6 Hz & 0.88 & 80.0 & 4.20 \\
 & & Full & $+$5.5 Hz & \textbf{0.91} & 57.7 & 4.00\\
\midrule
\multirow{8}{*}{Speed} & \multirow{4}{*}{Quick $\rightarrow$ Slow} & 256 & \textbf{$-$2.29 SPS} & 0.81 & \textbf{88.0} & 3.82\\
 & & 384 & $-$1.74 SPS & 0.84 & 80.8 & 3.82 \\
 & & 512 & $-$1.93 SPS & 0.86 & 80.8 & 4.48 \\
 & & Full & $-$1.34 SPS & \textbf{0.90} & 77.8 & 4.33 \\
\cmidrule{2-7}
 & \multirow{4}{*}{Slow $\rightarrow$ Quick} & 256 & \textbf{$+$1.02 SPS} & 0.86 & \textbf{92.0} & 3.87 \\
 & & 384 & $+$0.74 SPS & 0.87 & \textbf{92.0} & 4.00 \\
 & & 512 & $+$0.69 SPS & 0.89 & 76.0 & 3.79 \\
 & & Full & $+$0.54 SPS & \textbf{0.91} & 84.6 & 4.18 \\
\bottomrule
\end{tabular*}
\caption{Intra-utterance style transition results with KV-cache buffer size $k=48$. $\Delta$ Metric measures pitch (Hz) or speed (SPS) differences between the first and last 3-second segments. SIM, Trans., and TMOS denote speaker similarity, perceptual transition rate, and transition naturalness, respectively.}
\label{tab:intra_results}
\end{table*}

\paragraph{KV-Cache Swap.}
As the generation reaches the desired transition point $t^*$, we swap the initial KV-cache in Decoder-A with that from Decoder-B for \textit{initial $n$ positions} across all layers:
\begin{equation}
    \bm{K}^{(A)}_{1:n}, \bm{V}^{(A)}_{1:n} \leftarrow \bm{K}^{(B)}_{1:n}, \bm{V}^{(B)}_{1:n}.
\label{eq:kv_swap}
\end{equation}
We set $n = n_{\text{text}} + k$.
Here, $n_{\text{text}}$ is the number of tokens in the text prompt (the linguistic content to be spoken), and $k$ is an additional buffer that covers the early audio tokens where style characteristics are actively encoded during initial generation. 
This initial region is critical because, as discussed in our analysis, the decoder establishes style information not only through the prompt representation but also through these early generated tokens.

Simultaneously, we also replace the style representation used in cross-attention with the modified version $\bm{E}'$.
This ensures that both the initial KV-cache and the incoming style information reflect the target style.

\paragraph{Sliding-window Attention Masking.}
In addition, we discover that KV-cache swap alone is insufficient because standard self-attention allows the decoder to attend to all previously generated tokens. 
This means that even after replacing the initial region with target style KV-cache, the decoder can still access the original style information encoded in tokens from positions in-between ($n+1$ to $t^*$).
These intermediate tokens, already generated under the source style, would continue to influence subsequent generation and prevent effective style transition.

To address this issue, we introduce sliding-window attention masking~\cite{beltagy2020longformer,xiao2024sink,lee2026wand} that restricts self-attention to only two specific regions: (1) the initial $n$ positions containing target style KV-cache and (2) the most recent $w$ tokens in the local window.

Formally, let $i$ denote the current query position, $j$ denote the key position, $n$ denote the size of the replaced initial KV region, and $w$ denote the sliding-window size.
For a token at position $i > t^*$ (after the style transition point), the attention mask is defined as:
\begin{equation}
M_{ij} = \begin{cases}
    0 & \text{if } j \leq n \,\, \text{or} \,\, i - w \leq j \leq i \\
    -\infty & \text{otherwise}
\end{cases}
\label{eq:sliding_mask}
\end{equation}
where $M_{ij} = 0$ allows attention and $M_{ij} = -\infty$ blocks attention after softmax.

By attending only to the replaced initial region and recent tokens, the decoder gradually adopts the new style characteristics while maintaining local coherence.
Note that local coherence is crucial for the overall quality and naturalness of the generated speech~\cite{ye2025llasa}.
Tokens generated immediately after $t^*$ are influenced by both the new style (from the replaced initial region) and the local context (from the recent window), enabling smooth style progression rather than abrupt transitions.
See Appendix~\ref{appendix:algorithm} for the full algorithm.

\subsection{Experimental Results}

% \paragraph{Setup.}
% To generate sufficiently long speech for observing style transitions, we select 400 samples from the LibriTTS-R test set with text token lengths between 50 and 70.
% We compare different sliding-window sizes $w \in \{256, 384, 512, \text{Full}\}$ to analyze the effect of attention scope on style transition effectiveness and speech quality, where ``Full'' denotes standard self-attention without sliding-window masking.
% See Section~\ref{appendix:joint_effect} for the ablation study on the KV-cache size $k$.

\paragraph{Setup.}
We select 400 LibriTTS-R samples with text token lengths between 50 and 70 and, for consistency across samples, set $t^*$ near the generation midpoint (approximately 400 audio tokens).
We use interpolation strength $\alpha = 2.0$ to compute the target style representation $\bm{E}'$ for the dual-decoder setup.
We compare different sliding-window sizes $w \in \{256, 384, 512, \text{Full}\}$ to analyze the effect of attention scope on style transition effectiveness and speech quality, where ``Full'' denotes standard self-attention without sliding-window masking.

\paragraph{Objective Results.}
Table~\ref{tab:intra_results} presents the results for different window sizes and style attribute changes. 
As expected, smaller windows produce more pronounced changes: window size of 256 achieves $+27.4$ Hz for Low$\rightarrow$High pitch conversion (vs. $+5.5$ Hz for Full) and $-2.29$ SPS for Quick$\rightarrow$Slow speed conversion (vs. $-1.34$ SPS for Full).
However, speaker similarity decreases with smaller windows (0.81 for size 256 vs. 0.90--0.91 for Full), demonstrating a trade-off between transition strength and identity preservation.
We therefore recommend $w=384$--$512$ for balance, or $w=256$ when transition strength is prioritized.

\paragraph{Subjective Results.}
Subjective evaluations confirm this trade-off.
For the perceptual transition detection, window size of 256 achieves the highest rate (88.0--96.2\% across attributes), while Full attention showed lower detection rates (55.6--84.6\%).
For smoothness, all configurations achieved scores above 3.48, with larger windows (512, Full) reaching up to 4.48, implying that more gradual transitions appear more natural.

\paragraph{KV-Cache Buffer Ablation.}
To examine the role of the audio-token buffer, we ablate $k$ across window sizes (see Appendix~\ref{appendix:joint_effect} for details). For High$\rightarrow$Low pitch conversion, swapping only the text positions ($k=0$) instead increases pitch by $6.8$--$16.2$ Hz, opposite to the target direction. Including at least 32 initial audio-token positions reduces pitch by $5.3$--$12.4$ Hz, supporting our use of $n=n_{\text{text}}+k$ to include early audio tokens in the swapped region.

\paragraph{Transition Dynamics.}
To characterize how the transition develops over time, we measure F0 in 0.5-second bins for 400 utterances per condition. For Low$\rightarrow$High conversion with $w=256$, the F0 shift grows from approximately $+7$ Hz at 0.25 seconds after $t^*$ to $+27$ Hz at 4.25 seconds, reaching 80\% of the final change after 3.3 seconds. The full-attention condition yields a $+12$ Hz shift, while the condition without KV-cache swapping yields no change. In a matched comparison using approximately 95 prompts per direction, fixed-index and nearest-word-boundary switching produce similar F0 changes ($+30$ vs. $+33$ Hz) and speaker similarity (0.81 vs. 0.82). Within-word switching, however, shows local F0 jumps of 7--9 Hz; thus, these aggregate results do not rule out brief local phonetic artifacts. Further spectrogram and waveform analyses are provided in Appendix~\ref{appendix:spectrogram}.

\paragraph{Adaptive Window Schedule.}
In a separate test, using $w=256$ for the first 3 seconds after $t^*$ and then increasing it to 512 yields a $+22.7$ Hz Low$\rightarrow$High pitch change with SIM of 0.86, compared with $+30.5$ Hz and SIM of 0.80 when keeping $w=256$ throughout. Switching to full attention after 3 seconds yields a $+14.0$ Hz change and SIM of 0.90. For Quick$\rightarrow$Slow conversion, the adaptive schedule retains the measured speed change while increasing SIM from 0.81 to 0.88 (see Appendix~\ref{appendix:joint_effect}).

\paragraph{Inference Latency.}
We measure inference latency on a single NVIDIA V100 GPU by generating 100 sentences with text token lengths $\geq 50$. As shown in Table~\ref{tab:latency}, our dual-decoder method adds only $0.77$ seconds ($\approx 2\%$) over the baseline. This overhead is small because Decoder-B generates only $k$ audio tokens after prompt prefill, producing a cache over $n = n_{\text{text}} + k$ positions, while Decoder-A performs the full autoregressive generation.

\begin{table}[t]
    \centering
    \setlength{\tabcolsep}{9pt}
    \renewcommand{\arraystretch}{1.1}
    \small
    \begin{tabular}{lcc}
    \toprule
    \textbf{Method} & \textbf{Latency (s)} & \textbf{RTF} \\
    \midrule
    Baseline & 36.90 $\pm$ 7.76 & 2.56 \\
    Ours & 37.67 $\pm$ 8.34 & 2.67 \\
    \bottomrule
    \end{tabular}
    \caption{Inference latency on a single NVIDIA V100 GPU for 100 sentences with text token lengths $\geq 50$.}
    \label{tab:latency}
\end{table}

\begin{table}[t]
    \centering
    \renewcommand{\arraystretch}{1.1}
    \footnotesize
    \resizebox{\columnwidth}{!}{%
    \begin{tabular}{llcc}
    \toprule
    \textbf{Attribute} & \textbf{Direction} & \textbf{Style diff} & \textbf{SIM} \\
    \midrule
    \multirow{2}{*}{Pitch} & High $\rightarrow$ Low & $-$4.3 Hz & 0.92 \\
                           & Low $\rightarrow$ High & $-$2.40 Hz & 0.92 \\
    \midrule
    \multirow{2}{*}{Speed} & Quick $\rightarrow$ Slow & $-$0.23 SPS & 0.93 \\
                           & Slow $\rightarrow$ Quick & $-$0.27 SPS & 0.91 \\
    \bottomrule
    \end{tabular}
    }
    \caption{Style transition results when only replacing the style embedding without KV swap.}
    \label{tab:ablation_style_only}
\end{table}
\paragraph{Ablation: Style Replacement Without KV-Cache Swap.}~\label{analysis:ablation}
In Table~\ref{tab:ablation_style_only}, we investigate whether simply replacing the style representation mid-generation is sufficient for style transition, without any KV-cache swap.
The results show that replacing only the style representation produces minimal style change. 
This confirms our analysis in Section~\ref{sec:early_bias}; due to the early-token bias, the model continues to attend to the cached style information from the initial tokens, making the newly introduced style representation ineffective. 
This demonstrates that the proposed KV-cache swap is essential for achieving intra-utterance style transitions.

\subsection{Generalization to Different Architectures and Categorical Attributes}
\label{sec:cross_model}
\begin{table}[t]
    \centering
    \renewcommand{\arraystretch}{1.1}
    \setlength{\tabcolsep}{12pt}
    \footnotesize
    \resizebox{\columnwidth}{!}{%
    \begin{tabular}{lcc}
    \toprule
    \textbf{Transition} & \textbf{Change (\%)} & \textbf{SIM} \\
    \midrule
    Happy $\rightarrow$ Calm  & 78 & 0.90 \\
    Angry $\rightarrow$ Calm  & 87 & 0.88 \\
    Angry $\rightarrow$ Happy & 74 & 0.90 \\
    Happy $\rightarrow$ Angry & 80 & 0.88 \\
    Surprised $\rightarrow$ Calm & 71 & 0.92 \\
    \bottomrule
    \end{tabular}
    }
    \caption{Mid-utterance emotion transition results on IndexTTS~2.}
    \label{tab:indextts_emotion}
\end{table}
To test generalization beyond Parler-TTS, we apply our method to IndexTTS~2~\cite{zhou2026indextts2}, an autoregressive zero-shot TTS model with a different architecture.
Unlike Parler-TTS, which uses natural-language prompts for continuous style attributes, IndexTTS~2 is conditioned on categorical emotion labels such as \textit{Happy}, \textit{Angry}, and \textit{Calm}.
On 100 long prompts with $w=256$, we measure dominant emotion changes between the first and last 3-second segments using emotion2vec~\cite{ma2024emotion2vec}, along with speaker similarity.
Across the five emotion transitions in Table~\ref{tab:indextts_emotion}, change rates range from 71--87\% and speaker similarity ranges from 0.88 to 0.92, indicating that our method extends across architectures and conditioning formats.
\section{Conclusion}\label{sec:conclusion}
We presented training-free methods for inter-utterance style interpolation and intra-utterance style transitions in autoregressive TTS models.
Embedding-space direction vectors enable continuous global control, while KV-cache swap and sliding-window masking mitigate style self-referencing to support smooth within-utterance changes.
Experiments demonstrate effective style control while preserving speaker identity, with IndexTTS~2 results extending the findings across architectures and categorical attributes.
Overall, our findings show that direct inference-time manipulation of style representations and decoder attention can expand TTS controllability without retraining.

\section*{Limitations}
Our main experiments cover pitch, speed, and gender, while IndexTTS~2 (Section~\ref{sec:cross_model}) extends the study to categorical emotions; other attributes such as intonation and prosodic emphasis remain future work.
Window size trades transition strength against speaker similarity and naturalness, motivating methods that better balance these objectives.
Our structured prompts make attribute positions explicit, but automatically identifying them in free-form prompts remains future work.
While the additional memory and computational costs are minimal, intra-utterance style transition requires two decoder inferences; practicality could be improved through lightweight approaches.
Although we validate our methods on autoregressive style-conditioned TTS models with different conditioning formats, including natural-language style prompts and categorical emotion labels, their applicability to non-autoregressive or diffusion-based models~\cite{moon2026deltatts} that lack KV-cache structures remains an open question.
Finally, although this work focuses on training-free continuous style control, extending these ideas to training-based approaches remains a promising direction.

\section*{Ethics Statement}
While our proposed style control methods do not aim to mimic a real person's voice and focus on enhancing the controllability of speech synthesis, the domain requires ethical consideration regarding potential misuse.
Continuous style interpolation techniques could be exploited to imitate or manipulate a specific speaker's voice, potentially leading to fraud, disinformation, or non-consensual voice cloning.
To prevent such risks, detection technologies that distinguish synthetic speech from real speech, as well as mechanisms such as audio watermarking, should be developed in parallel.

\section*{Acknowledgments}
This work was partly supported by the National Research Foundation of Korea (NRF) grant funded by the Korea government (MSIT) (RS-2026-25602529). This work was also partly supported by the Institute of Information \& Communications Technology Planning \& Evaluation (IITP) grants funded by the Korea government (MSIT) (IITP-2026-RS-2026-25615334, AI Star Fellowship Support Program; RS-2019-II190421, Artificial Intelligence Graduate School Program (Sungkyunkwan University)). This work was also partly supported by the Institute of Information \& Communications Technology Planning \& Evaluation (IITP, AI Computing Support Project for R\&D) grant funded by the Korea government (MSIT) (High-Performance Research AI Computing Infrastructure Support at the 2 PFLOPS Scale, 25\%).

\bibliography{references}

@inproceedings{guo2023prompttts,
  title={PromptTTS: Controllable Text-to-Speech with Text Descriptions},
  author={Guo, Zhifang and Leng, Yichong and Wu, Yihan and Zhao, Sheng and Tan, Xuejiao},
  booktitle={ICASSP 2023-2023 IEEE International Conference on Acoustics, Speech and Signal Processing (ICASSP)},
  pages={1--5},
  year={2023},
  organization={IEEE}
}

@article{yang2024instructtts,
  title={{InstructTTS}: Modelling Expressive TTS in Discrete Latent Space With Natural Language Style Prompt},
  author={Yang, Dongchao and Liu, Songxiang and Huang, Rongjie and Weng, Chao and Meng, Helen},
  journal={IEEE/ACM Transactions on Audio, Speech, and Language Processing},
  volume={32},
  pages={2913--2925},
  year={2024},
  doi={10.1109/TASLP.2024.3402088}
}

@inproceedings{
    hsu2018hierarchical,
    title={Hierarchical Generative Modeling for Controllable Speech Synthesis},
    author={Wei-Ning Hsu and Yu Zhang and Ron Weiss and Heiga Zen and Yonghui Wu and Yuan Cao and Yuxuan Wang},
    booktitle={International Conference on Learning Representations},
    year={2019}
}

@inproceedings{zhang2019learning,
  title={Learning latent representations for style control and transfer in end-to-end speech synthesis},
  author={Zhang, Ya-Jie and Pan, Shifeng and He, Lei and Ling, Zhen-Hua},
  booktitle={ICASSP 2019-2019 IEEE International Conference on Acoustics, Speech and Signal Processing (ICASSP)},
  pages={6945--6949},
  year={2019},
  organization={IEEE}
}

@article{chen2024valle2,
  title={Vall-e 2: Neural codec language models are human parity zero-shot text to speech synthesizers},
  author={Chen, Sanyuan and Liu, Shujie and Zhou, Long and Liu, Yanqing and Tan, Xu and Li, Jinyu and Zhao, Sheng and Qian, Yao and Wei, Furu},
  journal={arXiv preprint arXiv:2406.05370},
  year={2024}
}

@inproceedings{ju2024naturalspeech,
  title={NaturalSpeech 3: Zero-Shot Speech Synthesis with Factorized Codec and Diffusion Models},
  author={Ju, Zeqian and Wang, Yuancheng and Shen, Kai and Tan, Xu and Xin, Detai and Yang, Dongchao and Liu, Eric and Leng, Yichong and Song, Kaitao and Tang, Siliang and others},
  booktitle={International Conference on Machine Learning},
  pages={22605--22623},
  year={2024},
  organization={PMLR}
}

@inproceedings{leng2024prompttts2,
  title={PromptTTS 2: Describing and Generating Voices with Text Prompt},
  author={Leng, Yichong and Guo, Zhifang and Shen, Kai and Ju, Zeqian and Tan, Xu and Liu, Eric and Liu, Yufei and Yang, Dongchao and Zhang, Leying and Song, Kaitao and He, Lei and Li, Xiang-Yang and Zhao, Sheng and Qin, Tao and Bian, Jiang},
  booktitle={International Conference on Learning Representations},
  year={2024}
}

@article{lyth2024natural,
  title={Natural language guidance of high-fidelity text-to-speech with synthetic annotations},
  author={Lyth, Dan and King, Simon},
  journal={arXiv preprint arXiv:2402.01912},
  year={2024}
}

@article{du2024cosyvoice,
  title={CosyVoice: A Scalable Multilingual Zero-shot Text-to-speech Synthesizer based on Supervised Semantic Tokens},
  author={Du, Zhihao and Chen, Qian and Zhang, Shiliang and Hu, Kai and Lu, Heng and Yang, Yexin and Hu, Hangrui and Zheng, Siqi and Gu, Yue and Ma, Ziyang and others},
  journal={CoRR},
  year={2024}
}

@article{du2025cosyvoice,
  title={Cosyvoice 3: Towards in-the-wild speech generation via scaling-up and post-training},
  author={Du, Zhihao and Gao, Changfeng and Wang, Yuxuan and Yu, Fan and Zhao, Tianyu and Wang, Hao and Lv, Xiang and Wang, Hui and Ni, Chongjia and Shi, Xian and others},
  journal={arXiv preprint arXiv:2505.17589},
  year={2025}
}

@inproceedings{ren2021fastspeech,
    title={FastSpeech 2: Fast and High-Quality End-to-End Text to Speech},
    author={Yi Ren and Chenxu Hu and Xu Tan and Tao Qin and Sheng Zhao and Zhou Zhao and Tie-Yan Liu},
    booktitle={International Conference on Learning Representations},
    year={2021}
}

@inproceedings{lancucki2021fastpitch,
  title={Fastpitch: Parallel text-to-speech with pitch prediction},
  author={{\L}a{\'n}cucki, Adrian},
  booktitle={ICASSP 2021-2021 IEEE International Conference on Acoustics, Speech and Signal Processing (ICASSP)},
  pages={6588--6592},
  year={2021},
  organization={IEEE}
}

@inproceedings{wang2025wordlevel,
    title={Word-Level Emotional Expression Control in Zero-Shot Text-to-Speech Synthesis},
    author={Wang, Tianrui and Wang, Haoyu and Ge, Meng and Gong, Cheng and Qiang, Chunyu and Ma, Ziyang and Huang, Zikang and Yang, Guanrou and Wang, Xiaobao and Chng, Eng-Siong and Chen, Xie and Wang, Longbiao and Dang, Jianwu},
    booktitle={Advances in Neural Information Processing Systems},
    volume={38},
    pages={163639--163667},
    year={2025},
    doi={10.52202/085713-4930}
}

@article{xie2025emosteer,
  title={{EmoSteer-TTS}: Fine-Grained and Training-Free Emotion-Controllable Text-to-Speech via Activation Steering},
  author={Xie, Tianxin and Yang, Shan and Li, Chenxing and Yu, Dong and Liu, Li},
  journal={arXiv preprint arXiv:2508.03543},
  year={2025}
}

@article{mcinnes2018umap,
  title={Umap: Uniform manifold approximation and projection for dimension reduction},
  author={McInnes, Leland and Healy, John and Melville, James},
  journal={arXiv preprint arXiv:1802.03426},
  year={2018}
}

@article{koizumi2023libritts,
  title={LibriTTS-R: A Restored Multi-Speaker Text-to-Speech Corpus},
  author={Koizumi, Yuma and Zen, Heiga and Karita, Shigeki and Ding, Yifan and Yatabe, Kohei and Morioka, Nobuyuki and Bacchiani, Michiel and Zhang, Yu and Han, Wei and Bapna, Ankur},
  journal={Interspeech 2023},
  year={2023},
  publisher={ISCA}
}

@misc{lacombe-etal-2024-parler-tts,
  author = {Yoach Lacombe and Vaibhav Srivastav and Sanchit Gandhi},
  title = {Parler-TTS},
  year = {2024},
  publisher = {GitHub},
  journal = {GitHub repository},
  howpublished = {\url{https://github.com/huggingface/parler-tts}}
}

@article{morrison2023cross,
  title={Cross-domain neural pitch and periodicity estimation},
  author={Morrison, Max and Hsieh, Caedon and Pruyne, Nathan and Pardo, Bryan},
  journal={arXiv preprint arXiv:2301.12258},
  year={2023}
}

@article{chen2022wavlm,
  title={Wavlm: Large-scale self-supervised pre-training for full stack speech processing},
  author={Chen, Sanyuan and Wang, Chengyi and Chen, Zhengyang and Wu, Yu and Liu, Shujie and Chen, Zhuo and Li, Jinyu and Kanda, Naoyuki and Yoshioka, Takuya and Xiao, Xiong and others},
  journal={IEEE Journal of Selected Topics in Signal Processing},
  volume={16},
  number={6},
  pages={1505--1518},
  year={2022},
  publisher={IEEE}
}

@inproceedings{ji2024textrolspeech,
  title={Textrolspeech: A text style control speech corpus with codec language text-to-speech models},
  author={Ji, Shengpeng and Zuo, Jialong and Fang, Minghui and Jiang, Ziyue and Chen, Feiyang and Duan, Xinyu and Huai, Baoxing and Zhao, Zhou},
  booktitle={ICASSP 2024-2024 IEEE International Conference on Acoustics, Speech and Signal Processing (ICASSP)},
  pages={10301--10305},
  year={2024},
  organization={IEEE}
}

@inproceedings{korotkova2024word,
  title={Word-level text markup for prosody control in speech synthesis},
  author={Korotkova, Yuliya and Kalinovskiy, Ilya and Vakhrusheva, Tatiana},
  booktitle={Proc. Interspeech 2024},
  pages={2280--2284},
  year={2024}
}

@inproceedings{ji2025controlspeech,
  title={ControlSpeech: Towards simultaneous and independent zero-shot speaker cloning and zero-shot language style control},
  author={Ji, Shengpeng and Chen, Qian and Wang, Wen and Zuo, Jialong and Fang, Minghui and Jiang, Ziyue and Huang, Hai and Wang, Zehan and Cheng, Xize and Zheng, Siqi and others},
  booktitle={Proceedings of the 63rd Annual Meeting of the Association for Computational Linguistics (Volume 1: Long Papers)},
  pages={6966--6981},
  year={2025}
}

@article{beltagy2020longformer,
  title={Longformer: The long-document transformer},
  author={Beltagy, Iz and Peters, Matthew E and Cohan, Arman},
  journal={arXiv preprint arXiv:2004.05150},
  year={2020}
}

@inproceedings{guo2024audiobook,
  title     = {{Text-aware and Context-aware Expressive Audiobook Speech Synthesis}},
  author    = {Dake Guo and Xinfa Zhu and Liumeng Xue and Yongmao Zhang and Wenjie Tian and Lei Xie},
  year      = {2024},
  booktitle = {{Interspeech 2024}},
  pages     = {1790--1794},
  issn      = {2958-1796},
}

@inproceedings{liu2024emotion,
  title={Emotion rendering for conversational speech synthesis with heterogeneous graph-based context modeling},
  author={Liu, Rui and Hu, Yifan and Ren, Yi and Yin, Xiang and Li, Haizhou},
  booktitle={Proceedings of the AAAI Conference on Artificial Intelligence},
  volume={38},
  pages={18698--18706},
  year={2024}
}

@article{wang2025spark,
  title={Spark-tts: An efficient llm-based text-to-speech model with single-stream decoupled speech tokens},
  author={Wang, Xinsheng and Jiang, Mingqi and Ma, Ziyang and Zhang, Ziyu and Liu, Songxiang and Li, Linqin and Liang, Zheng and Zheng, Qixi and Wang, Rui and Feng, Xiaoqin and others},
  journal={arXiv preprint arXiv:2503.01710},
  year={2025}
}

@inproceedings{
xiao2024sink,
title={Efficient Streaming Language Models with Attention Sinks},
author={Guangxuan Xiao and Yuandong Tian and Beidi Chen and Song Han and Mike Lewis},
booktitle={The Twelfth International Conference on Learning Representations},
year={2024}
}

@inproceedings{wang2018style,
  title={Style tokens: Unsupervised style modeling, control and transfer in end-to-end speech synthesis},
  author={Wang, Yuxuan and Stanton, Daisy and Zhang, Yu and Ryan, RJ-Skerry and Battenberg, Eric and Shor, Joel and Xiao, Ying and Jia, Ye and Ren, Fei and Saurous, Rif A},
  booktitle={International conference on machine learning},
  pages={5180--5189},
  year={2018},
  organization={PMLR}
}

@article{wang2023neural,
  title={Neural codec language models are zero-shot text to speech synthesizers},
  author={Wang, Chengyi and Chen, Sanyuan and Wu, Yu and Zhang, Ziqiang and Zhou, Long and Liu, Shujie and Chen, Zhuo and Liu, Yanqing and Wang, Huaming and Li, Jinyu and others},
  journal={arXiv preprint arXiv:2301.02111},
  year={2023}
}

@article{kanda2024making,
  title={Making flow-matching-based zero-shot text-to-speech laugh as you like},
  author={Kanda, Naoyuki and Wang, Xiaofei and Eskimez, Sefik Emre and Thakker, Manthan and Yang, Hemin and Zhu, Zirun and Tang, Min and Li, Canrun and Tsai, Chung-Hsien and Xiao, Zhen and others},
  journal={arXiv preprint arXiv:2402.07383},
  year={2024}
}

@inproceedings{wu2024laugh,
  title={Laugh now cry later: Controlling time-varying emotional states of flow-matching-based zero-shot text-to-speech},
  author={Wu, Haibin and Wang, Xiaofei and Eskimez, Sefik Emre and Thakker, Manthan and Tompkins, Daniel and Tsai, Chung-Hsien and Li, Canrun and Xiao, Zhen and Zhao, Sheng and Li, Jinyu and others},
  booktitle={2024 IEEE Spoken Language Technology Workshop (SLT)},
  pages={690--697},
  year={2024},
  organization={IEEE}
}

@article{gao2025prompt,
  title={Prompt-Unseen-Emotion: Zero-shot Expressive Speech Synthesis with Prompt-LLM Contextual Knowledge for Mixed Emotions},
  author={Gao, Xiaoxue and Zhang, Huayun and Chen, Nancy F},
  journal={arXiv preprint arXiv:2506.02742},
  year={2025}
}

@inproceedings{lemerle2025lina,
  title={Lina-Style: Word-Level Style Control in TTS via Interleaved Synthetic Data},
  author={Lemerle, Th{\'e}odor and Obin, Nicolas and Roebel, Axel},
  booktitle={Proc. SSW 2025},
  pages={35--39},
  year={2025}
}

@article{ye2025llasa,
  title={{Llasa}: Scaling train-time and inference-time compute for llama-based speech synthesis},
  author={Ye, Zhen and Zhu, Xinfa and Chan, Chi-Min and Wang, Xinsheng and Tan, Xu and Lei, Jiahe and Peng, Yi and Liu, Haohe and Jin, Yizhu and Dai, Zheqi and others},
  journal={arXiv preprint arXiv:2502.04128},
  year={2025}
}

@inproceedings{zhou2026indextts2,
  title={IndexTTS2: A Breakthrough in Emotionally Expressive and Duration-Controlled Auto-Regressive Zero-Shot Text-to-Speech},
  author={Zhou, Siyi and Zhou, Yiquan and He, Yi and Zhou, Xun and Wang, Jinchao and Deng, Wei and Shu, Jingchen},
  booktitle={Proceedings of the AAAI Conference on Artificial Intelligence},
  volume={40},
  pages={35139--35148},
  year={2026}
}

@inproceedings{ma2024emotion2vec,
  title={emotion2vec: Self-supervised pre-training for speech emotion representation},
  author={Ma, Ziyang and Zheng, Zhisheng and Ye, Jiaxin and Li, Jinchao and Gao, Zhifu and Zhang, Shiliang and Chen, Xie},
  booktitle={Findings of the ACL 2024},
  pages={15747--15760},
  year={2024}
}

@article{zou2023representation,
  title={Representation engineering: A top-down approach to ai transparency},
  author={Zou, Andy and Phan, Long and Chen, Sarah and Campbell, James and Guo, Phillip and Ren, Richard and Pan, Alexander and Yin, Xuwang and Mazeika, Mantas and Dombrowski, Ann-Kathrin and others},
  journal={arXiv preprint arXiv:2310.01405},
  year={2023}
}

@article{panickssery2023steering,
  title={Steering llama 2 via contrastive activation addition},
  author={Panickssery, Nina and Gabrieli, Nick and Schulz, Julian and Tong, Meg and Hubinger, Evan and Turner, Alexander Matt},
  journal={arXiv preprint arXiv:2312.06681},
  year={2023}
}

@inproceedings{park2024linear,
  title={The Linear Representation Hypothesis and the Geometry of Large Language Models},
  author={Park, Kiho and Choe, Yo Joong and Veitch, Victor},
  booktitle={Proceedings of the 41st International Conference on Machine Learning},
  pages={39643--39666},
  year={2024},
  volume={235},
  series={Proceedings of Machine Learning Research},
  publisher={PMLR}
}

@article{gu2024attention,
  title={When attention sink emerges in language models: An empirical view},
  author={Gu, Xiangming and Pang, Tianyu and Du, Chao and Liu, Qian and Zhang, Fengzhuo and Du, Cunxiao and Wang, Ye and Lin, Min},
  journal={arXiv preprint arXiv:2410.10781},
  year={2024}
}

@article{barbero2025llms,
  title={Why do LLMs attend to the first token?},
  author={Barbero, Federico and Arroyo, Alvaro and Gu, Xiangming and Perivolaropoulos, Christos and Bronstein, Michael and Veli{\v{c}}kovi{\'c}, Petar and Pascanu, Razvan},
  journal={arXiv preprint arXiv:2504.02732},
  year={2025}
}

@inproceedings{chen2022fine,
  title={Fine-grained style control in transformer-based text-to-speech synthesis},
  author={Chen, Li-Wei and Rudnicky, Alexander},
  booktitle={ICASSP 2022-2022 IEEE International Conference on Acoustics, Speech and Signal Processing (ICASSP)},
  pages={7907--7911},
  year={2022},
  organization={IEEE}
}

@inproceedings{xie2025towards,
  title={Towards controllable speech synthesis in the era of large language models: A systematic survey},
  author={Xie, Tianxin and Rong, Yan and Zhang, Pengfei and Wang, Wenwu and Liu, Li},
  booktitle={Proceedings of the 2025 Conference on Empirical Methods in Natural Language Processing},
  pages={764--791},
  year={2025}
}

@article{liu2023promptstyle,
  title={Promptstyle: Controllable style transfer for text-to-speech with natural language descriptions},
  author={Liu, Guanghou and Zhang, Yongmao and Lei, Yi and Chen, Yunlin and Wang, Rui and Li, Zhifei and Xie, Lei},
  journal={arXiv preprint arXiv:2305.19522},
  year={2023}
}

@article{huang2025instructttseval,
  title={Instructttseval: Benchmarking complex natural-language instruction following in text-to-speech systems},
  author={Huang, Kexin and Tu, Qian and Fan, Liwei and Yang, Chenchen and Zhang, Dong and Li, Shimin and Fei, Zhaoye and Cheng, Qinyuan and Qiu, Xipeng},
  journal={arXiv preprint arXiv:2506.16381},
  year={2025}
}

@inproceedings{wang2023fine,
  title={Fine-grained emotional control of text-to-speech: Learning to rank inter-and intra-class emotion intensities},
  author={Wang, Shijun and Gu{\dh}nason, J{\'o}n and Borth, Damian},
  booktitle={ICASSP 2023-2023 IEEE International Conference on Acoustics, Speech and Signal Processing (ICASSP)},
  pages={1--5},
  year={2023},
  organization={IEEE}
}

@article{chen2026flexivoice,
  title={FlexiVoice: Enabling Flexible Style Control in Zero-Shot TTS with Natural Language Instructions},
  author={Chen, Dekun and Zhang, Xueyao and Wang, Yuancheng and Dai, Kenan and Ma, Li and Wu, Zhizheng},
  journal={arXiv preprint arXiv:2601.04656},
  year={2026}
}

@inproceedings{qiang2026instructaudio,
  title={Instructaudio: Unified speech and music generation with natural language instruction},
  author={Qiang, Chunyu and Yin, Kang and Wang, Xiaopeng and Liang, Yuzhe and Zhao, Jiahui and Fu, Ruibo and Wang, Tianrui and Gong, Cheng and Zhang, Chen and Wang, Longbiao and others},
  booktitle={ICASSP 2026-2026 IEEE International Conference on Acoustics, Speech and Signal Processing (ICASSP)},
  pages={17722--17726},
  year={2026},
  organization={IEEE}
}

@inproceedings{chen2026isse,
  title={ISSE: An Instruction-Guided Speech Style Editing Dataset and Benchmark},
  author={Chen, Yun and Chen, Qi and Dai, Zheqi and Singh, Arshdeep and Jackson, Philip JB and Plumbley, Mark D},
  booktitle={ICASSP 2026-2026 IEEE International Conference on Acoustics, Speech and Signal Processing (ICASSP)},
  pages={19482--19486},
  year={2026},
  organization={IEEE}
}

@article{wang2025controllable,
  title={Controllable Text-to-Speech Synthesis with Masked-Autoencoded Style-Rich Representation},
  author={Wang, Yongqi and Zhang, Chunlei and Chen, Hangting and Zhao, Zhou and Yu, Dong},
  journal={arXiv preprint arXiv:2506.02997},
  year={2025}
}

@inproceedings{kang2026encoderfree,
  title={Encoder-Free Style-Controllable Text-to-Speech with Voice Attribute Vectors},
  author={Kang, Jaehoon and Lee, Yejin and Shim, Kyuhong},
  booktitle={2026 International Conference on Electronics, Information, and Communication, ICEIC 2026},
  year={2026},
  doi={10.1109/ICEIC69189.2026.11386084}
}

@inproceedings{lee2026wand,
  title={{WAND}: Windowed Attention and Knowledge Distillation for Efficient Autoregressive Text-to-Speech Models},
  author={Lee, Hanna and Nguyen, Tan Dat and Kang, Jaehoon and Shim, Kyuhong},
  booktitle={Proceedings of Interspeech 2026},
  year={2026}
}

@inproceedings{moon2026deltatts,
  title={{DELTA-TTS}: Adapting Autoregressive Model into Diffusion Language Model for Text-to-Speech},
  author={Moon, Junwon and Lee, Yejin and Kim, Seungbeom and Ahn, Hoseong and Park, Sewoong and Kim, Heeseung and Shim, Kyuhong},
  booktitle={Advances in Neural Information Processing Systems},
  year={2026}
}

\newpage
\appendix
\section{Experimental Details}
\label{app:exp_details}

\subsection{Objective Evaluation Metrics}
\label{app:obj_metrics}

We use the following metrics for objective evaluation across both inter-utterance and intra-utterance experiments. \textbf{Pitch.} We measure the average fundamental frequency (F0) in Hz using PENN (Pitch Estimator Neural Network)~\cite{morrison2023cross}. \textbf{Speed.} We compute speaking rate as syllables per second (SPS): the total number of syllables divided by the utterance duration~\cite{wang2025spark}. \textbf{Speaker Similarity.} We extract speaker embeddings using a fine-tuned WavLM\footnote{microsoft/wavlm-base-plus-sv}~\cite{chen2022wavlm} and compute cosine similarity between audio segments to verify that speaker identity is preserved. \textbf{Gender Classification.} For gender conversion evaluation, we use a pre-trained gender classification model\footnote{alefiury/wav2vec2-large-xlsr-53-gender-recognition-librispeech} to measure the conversion success rate.

\subsection{Dataset and Model}
\label{app:dataset}

For inter-utterance style interpolation, we selected 400 sentences from the LibriTTS-R~\cite{koizumi2023libritts} test-clean split.
For intra-utterance style variation, we selected 400 samples from the LibriTTS-R test-clean split where the text token length is between 50 and 70 to ensure sufficiently long speech for observing style transitions.
All audio samples were generated using the Parler-TTS-mini model~\cite{lacombe-etal-2024-parler-tts,lyth2024natural}.

\subsection{Inter-Utterance Style Interpolation}
\label{app:inter_details}

\paragraph{Objective Evaluation.}
We generate speech at various interpolation strengths and measure the change compared to the original style ($\alpha=0$).
For gender conversion, we evaluate $\alpha\in\{0.5, 1.0, 1.5, 2.0\}$, while for pitch and speed conversion, we extend the range to $\alpha\in\{-1.0, -0.5, 0.5, 1.0, 1.5, 2.0\}$ to examine both interpolation and extrapolation.
We use the metrics described in Appendix~\ref{app:obj_metrics}.

\begin{figure}[t]
    \centering
    \includegraphics[width=\columnwidth]{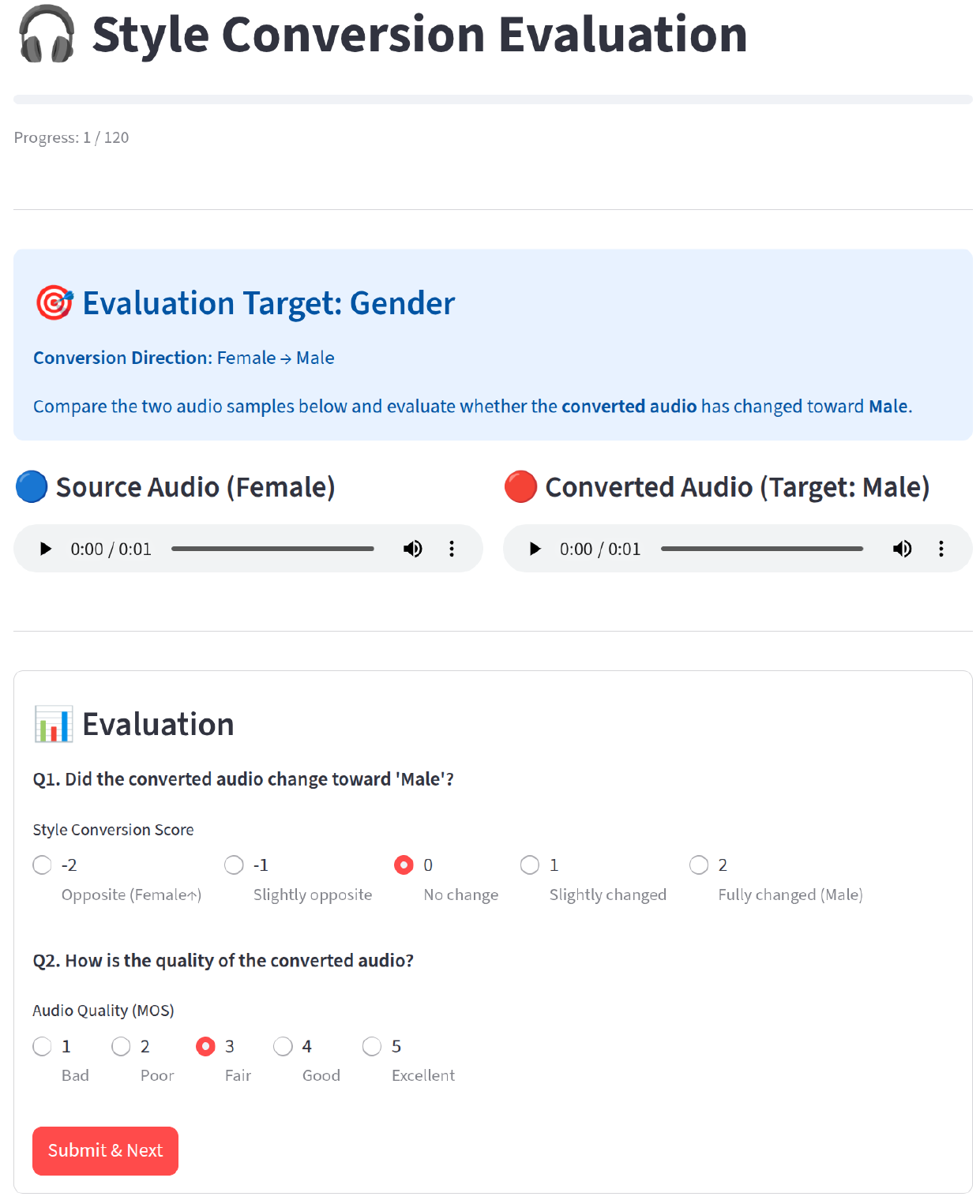}
    \caption{Evaluation interface for inter-utterance style interpolation. Participants compared source and converted audio, rating both the style conversion effectiveness (Q1) and audio quality (Q2).}
    \label{fig:inter_interface}
\end{figure}

\paragraph{Subjective Evaluation.}
We recruited 15 participants to evaluate the converted speech across different interpolation strengths $\alpha \in \{0.5, 1.0, 2.0\}$. Figure~\ref{fig:inter_interface} shows the evaluation interface. For each sample, participants were presented with two audio clips: the source audio and the converted audio. The evaluation target (Gender, Pitch, or Speed) and conversion direction were clearly indicated. Participants could listen to both audio samples multiple times before rating two criteria. First, they evaluated the \textbf{Style Conversion Score}, indicating whether the converted audio changed toward the target style on a 5-point scale: $-2$ (opposite direction), $-1$ (slightly opposite), $0$ (no change), $+1$ (slightly changed), and $+2$ (fully changed). Second, they rated the \textbf{Audio Quality (MOS)} on a 5-point scale ranging from 1 (Bad) to 5 (Excellent).

\begin{figure}[t]
    \centering
    \includegraphics[width=\columnwidth]{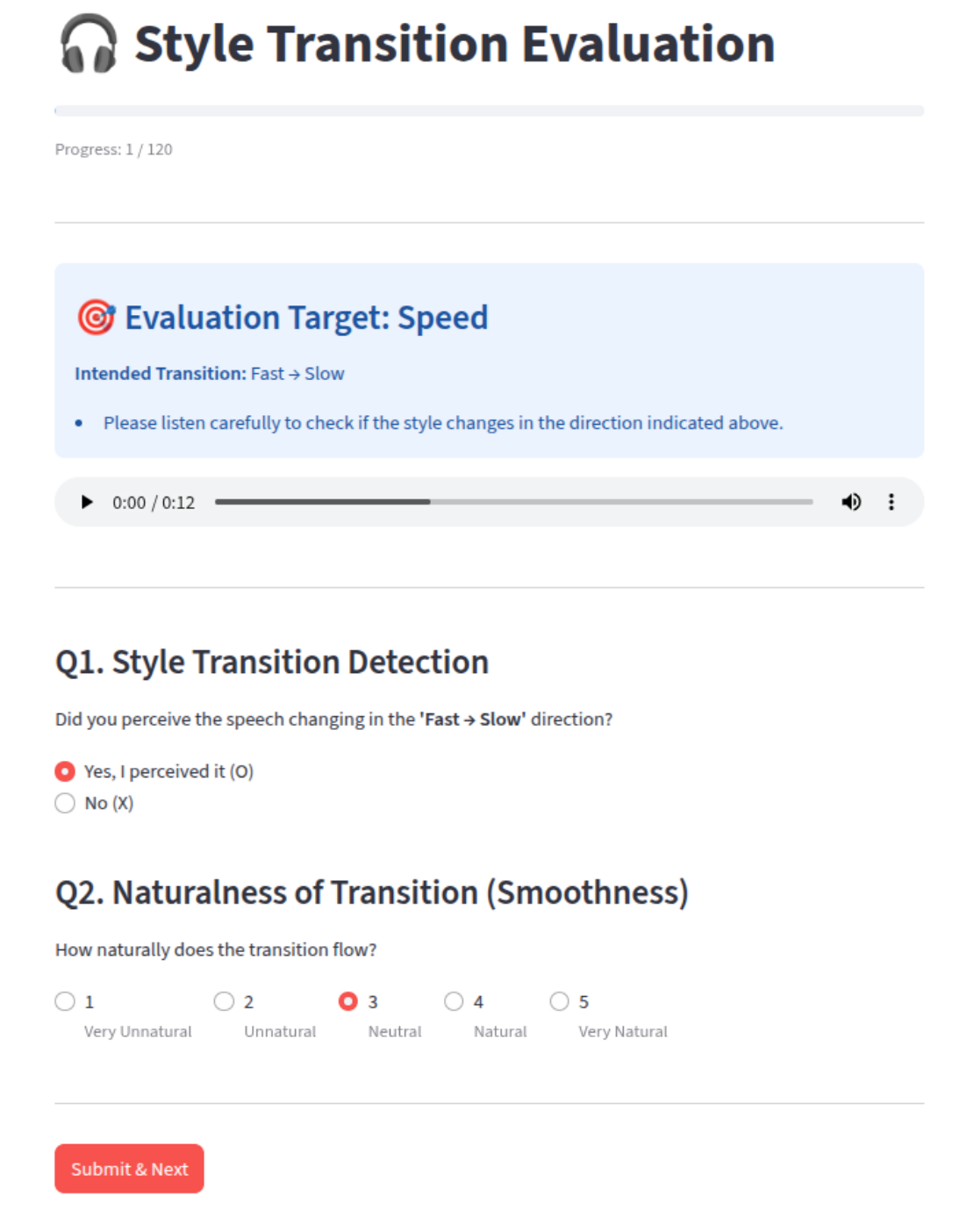}
    \caption{Evaluation interface for intra-utterance style transition. Participants evaluated whether they detected the intended style transition (Q1) and rated the naturalness of the transition (Q2).}
    \label{fig:intra_interface}
\end{figure}

\subsection{Intra-Utterance Style Transition}
\label{app:intra_details}

\paragraph{Objective Evaluation.}
To measure style transition effectiveness, we extract the first 3 seconds and the last 3 seconds of each generated utterance and compute the difference in style attributes between these segments using the metrics in Appendix~\ref{app:obj_metrics}. Specifically, we calculate the \textbf{Pitch diff} (difference in average F0) and \textbf{Speed diff} (difference in speaking rate) between the two segments. Additionally, we compute \textbf{Speaker Similarity (SIM)} using cosine similarity between speaker embeddings to verify that speaker identity is preserved across the transition.

\paragraph{Subjective Evaluation.}
We evaluated both the detectability and smoothness of style transitions using the interface shown in Figure~\ref{fig:intra_interface}. For each sample, participants listened to a single audio clip containing the style transition, with the intended target (e.g., Speed) and direction (e.g., Quick $\rightarrow$ Slow) displayed. Participants answered two questions: \textbf{Q1 (Style Transition Detection)}, a binary choice of whether they perceived the speech changing in the intended direction; and \textbf{Q2 (Smoothness, TMOS)}, rating the naturalness of the transition on a 5-point scale from 1 (Very Unnatural) to 5 (Very Natural). The detection rate is reported as Trans.~(\%) in Table~\ref{tab:intra_results}.

\section{Full Vector vs. Attribute-Only Vector Interpolation}
\label{appendix:full_vs_attribute}

In Section~\ref{sec:inter_utterance}, we apply the direction vector only to attribute token positions $i \in \mathcal{A}$, leaving other token embeddings unchanged. 
An alternative approach is to shift all token positions toward the target style representation.

We compare these two strategies by conducting experiments where all tokens are shifted. 
For attribute tokens, we vary $\alpha \in \{-1.0, -0.5, 0.5, 1.0, 1.5, 2.0\}$, while for non-attribute tokens, we apply a separate interpolation factor $\beta \in \{-0.5, -0.25, 0, 0.25, 0.5, 0.75, 1.0\}$ with increments of 0.25.

\begin{figure*}[t]
\centering
\includegraphics[width=\textwidth]{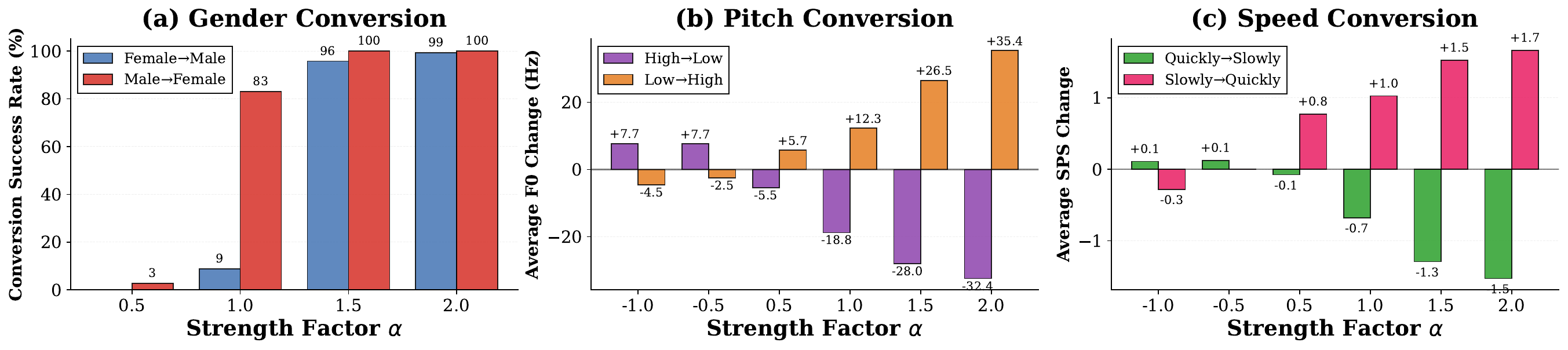}
\caption{Inter-utterance style interpolation results when shifting all tokens. Results are comparable to attribute-only interpolation (Figure~\ref{fig:inter_utterance}), demonstrating that manipulating only attribute tokens is sufficient for effective style control.}
\label{fig:full_vector_interpolation}
\end{figure*}

Figure~\ref{fig:full_vector_interpolation} presents the results when shifting all tokens. 
Compared to the attribute-only results in Figure~\ref{fig:inter_utterance}, both approaches achieve similar performance across all three attributes: gender conversion reaches a 99--100\% success rate at $\alpha = 2.0$, pitch changes reach up to $-$32.4 Hz (High$\rightarrow$Low) and $+$35.4 Hz (Low$\rightarrow$High), and speed changes reach up to $-$1.5 SPS (Fast$\rightarrow$Slow) and $+$1.7 SPS (Slow$\rightarrow$Fast).

Table~\ref{tab:full_vector_results} summarizes the results at $\alpha = 2.0$ and $\beta = 1.0$. 
Compared to the attribute-only approach (Table~\ref{tab:inter_utterance}), the full vector interpolation achieves comparable success rates and acoustic feature changes while maintaining similar speaker similarity. 
Since the full vector approach requires additional hyperparameter tuning for non-attribute tokens without providing clear benefits, we adopt the simpler attribute-only strategy in our main experiments.

\begin{table}[t!]
\centering
\small
\resizebox{\columnwidth}{!}{%
\begin{tabular}{llccc}
\toprule
\textbf{Attribute} & \textbf{Direction} & \textbf{Success (\%)} & \textbf{$\Delta$ Metric} & \textbf{SIM} \\
\midrule
\multirow{2}{*}{Gender} & Female $\rightarrow$ Male & 99.3 & -- & -- \\
                        & Male $\rightarrow$ Female & 100 & -- & -- \\
\midrule
\multirow{2}{*}{Pitch}  & High $\rightarrow$ Low & 91.3 & $-32.4$ Hz & 0.79 \\
                        & Low $\rightarrow$ High & 93.5 & $+35.4$ Hz & 0.79 \\
\midrule
\multirow{2}{*}{Speed}  & Fast $\rightarrow$ Slow & 95.5 & $-1.5$ SPS & 0.85 \\
                        & Slow $\rightarrow$ Fast & 95.75 & $+1.7$ SPS & 0.85 \\
\bottomrule
\end{tabular}%
}
\caption{Full vector interpolation results at $\alpha = 2.0$ (transition strength for attribute tokens) and $\beta = 1.0$ (transition strength for non-attribute tokens).}
\label{tab:full_vector_results}
\end{table}

\section{Extrapolation Beyond the Source--Target Interval}
\label{appendix:extrapolation}

We further analyze the behavior of style-vector extrapolation outside the source--target interval.
Using a fixed subset of 100 sentences from the 400-sentence LibriTTS-R test-clean set, we generate speech for six conversion directions (gender, pitch, and speed in both directions) with
$\alpha \in \{-3,-2,-1,0,1,2,3,4,5,7,10\}$, yielding 6,600 utterances in total.
We evaluate style changes using the attribute-specific metrics in Appendix~\ref{app:obj_metrics} and report the generation failure rate.
We define failures as silent, severely truncated, excessively long, or unvoiced outputs.
Attribute values are averaged over valid outputs, whereas failure rates are computed over all 100 samples in each configuration.

Table~\ref{tab:extrapolation_summary} summarizes moderate positive extrapolation at $\alpha=3$.
Across all directions, the target styles become stronger or saturated while failure rates remain at or below 2\%.
Beyond $\alpha=3$, the style effect generally saturates or becomes non-monotonic, indicating that increasing $\alpha$ does not provide unbounded amplification.

\begin{table}[t]
\centering
\small
\resizebox{\columnwidth}{!}{%
\begin{tabular}{llcc}
\toprule
\textbf{Attribute} & \textbf{Direction} & $\boldsymbol{\alpha=3}$ & \textbf{Gen. Failure (\%)} \\
\midrule
\multirow{2}{*}{Gender} & F $\rightarrow$ M & 98.0\% & $\leq 2$ \\
                        & M $\rightarrow$ F & 96.7\% & $\leq 2$ \\
\midrule
\multirow{2}{*}{Pitch}  & H $\rightarrow$ L & $-36.1$ Hz & $\leq 2$ \\
                        & L $\rightarrow$ H & $+42.8$ Hz & $\leq 2$ \\
\midrule
\multirow{2}{*}{Speed}  & Q $\rightarrow$ S & $-2.0$ SPS & $\leq 2$ \\
                        & S $\rightarrow$ Q & $+1.7$ SPS & $\leq 2$ \\
\bottomrule
\end{tabular}%
}
\caption{Summary of positive extrapolation at $\alpha=3$. Attribute values are computed over valid outputs, while failure rates are computed over all samples.}
\label{tab:extrapolation_summary}
\end{table}

At larger positive $\alpha$, stylistic intensification becomes inconsistent and generation failures become more frequent.
The observed failures are primarily severely truncated or excessively long outputs.

Negative extrapolation is substantially less stable and strongly direction-dependent.
At $\alpha=-2$, Female$\rightarrow$Male and Low$\rightarrow$High pitch reach failure rates of 59\% and 32\%, respectively, whereas High$\rightarrow$Low pitch and both speed directions remain largely stable even at $\alpha=-3$.
The range used in our main experiments, including $\alpha=-1$, remains stable across directions, with failure rates no greater than 8\%.
Overall, these results show that the controllable embedding region is bounded and asymmetric: moderate extrapolation can strengthen some attributes, whereas aggressive extrapolation leads to saturation, non-monotonic behavior, and eventually generation failure.

\section{Quantitative Analysis of Early-Token Attention Bias}
\label{appendix:attention_variance}

To quantitatively support the observation in Section~\ref{sec:early_bias} that cross-attention weights stabilize after the initial generation phase, we compute the variance of attention weights across style tokens at each audio generation position.

Specifically, let $\mathbf{a}_t \in \mathbb{R}^{|S|}$ denote the attention weight distribution over style tokens $S$ at audio position $t$. We compute the variance of this distribution:
\begin{equation}
\text{Var}(t) = \frac{1}{|S|} \sum_{s \in S} \left( a_{t,s} - \bar{a}_t \right)^2
\end{equation}
where $a_{t,s}$ is the attention weight from audio position $t$ to style token $s$, and $\bar{a}_t$ is the mean attention weight at position $t$.

\begin{figure}[t]
\centering
\includegraphics[width=0.85\linewidth]{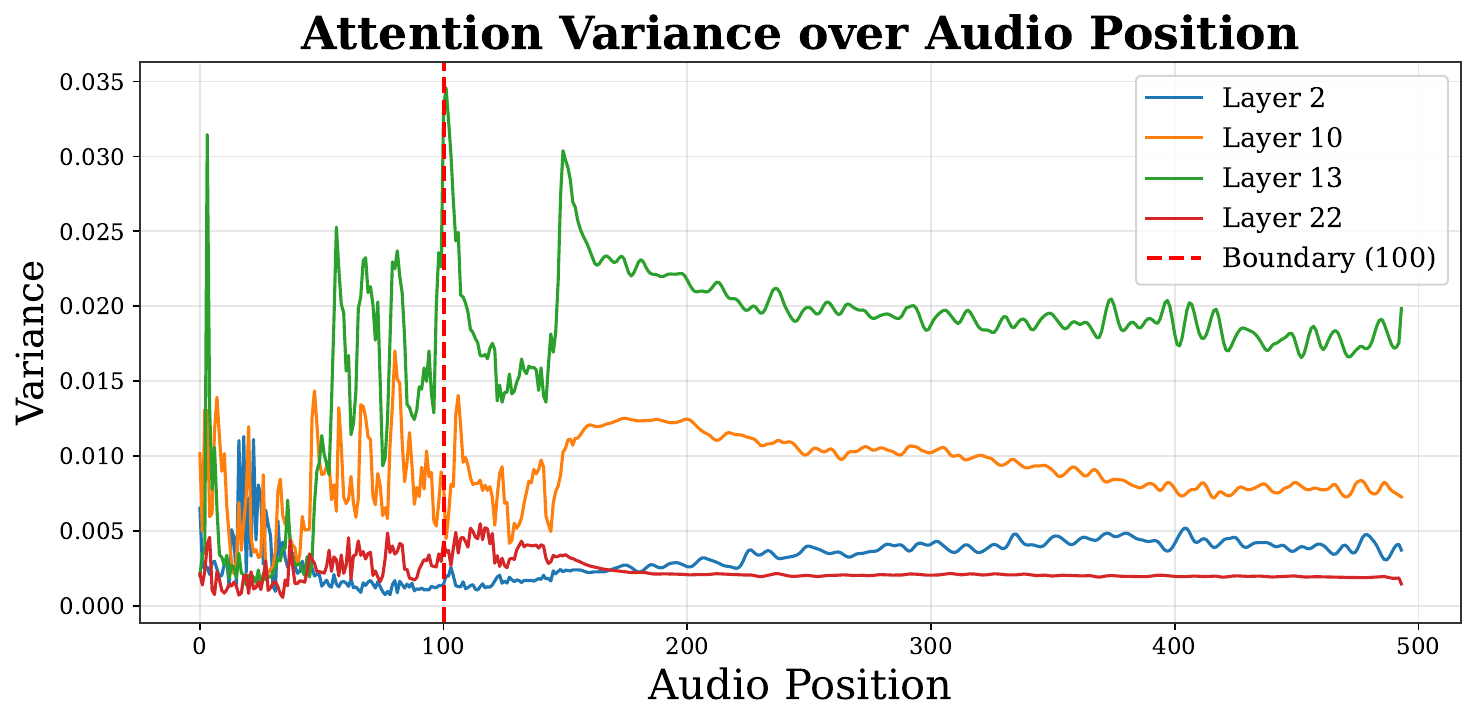}
\caption{Attention weight variance across audio generation positions for different decoder layers. During early generation, variance fluctuates as the model actively updates attention weights to establish style characteristics. Afterward, variance becomes stable, indicating fixed attention patterns.}
\label{fig:attention_variance}
\end{figure}

Figure~\ref{fig:attention_variance} shows the attention variance across audio positions for multiple decoder layers. 
During the initial generation phase, the variance fluctuates significantly, indicating active attention weight updates as the model establishes style characteristics. 
After this phase, the variance stabilizes, confirming that fixed attention weights are assigned across all style tokens. 
This quantitative analysis supports our claim that the decoder actively queries style information during early generation and subsequently maintains a static attention pattern.

\section{Style Prompt Examples}
\label{appendix:prompts}

Table~\ref{tab:style_prompts} lists the style prompts used in our experiments for inter-utterance style interpolation and intra-utterance style variation. For each attribute, we define source and target prompts that differ only in the attribute-specific keywords (underlined).

\begin{table}[t!]
\centering
\vspace{0.2cm}
\small
\begin{tabular}{@{}lp{5.8cm}@{}}
\toprule
\textbf{Attribute} & \textbf{Prompt} \\
\midrule
\multirow{2}{*}{Gender} & ``A \underline{male} voice speaks moderate at a medium pitch with monotone modulation and a clean quality.'' \\
                        & ``A \underline{female} voice speaks moderate at a medium pitch with monotone modulation and a clean quality.'' \\
\midrule
\multirow{2}{*}{Pitch}  & ``A male voice speaks normally at a \underline{high} pitch and a clean quality.'' \\
                        & ``A male voice speaks normally at a \underline{low} pitch and a clean quality.'' \\
\midrule
\multirow{2}{*}{Speed}  & ``A male voice speaks \underline{quickly} at a normal pitch and a clean quality.'' \\
                        & ``A male voice speaks \underline{slowly} at a normal pitch and a clean quality.'' \\
\bottomrule
\end{tabular}
\caption{Style prompts used in experiments. The attribute keywords (underlined) are the tokens where direction vectors are computed and applied.}
\label{tab:style_prompts}
\end{table}

For intra-utterance experiments, we use the same prompts but apply the direction vector manipulation at the transition point during autoregressive generation.

\section{Joint Effect of Window and KV-Cache Size}
\label{appendix:joint_effect}

Our method has two key hyperparameters: the sliding-window size $w$ and the KV-cache region size $n = n_{\text{text}} + k$. 
We conduct a grid search over window sizes ($w \in \{256, 384, 512\}$) and additional KV-cache tokens ($k \in \{0, 32, 48\}$) to analyze their joint effect on style transition.

\begin{figure}[t]
    \centering
    \includegraphics[width=0.8\linewidth]{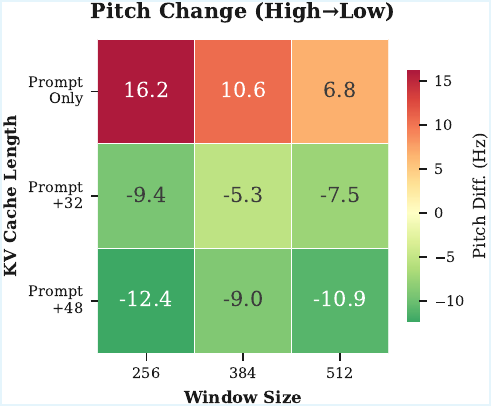}
    \caption{Ablation study on window size and KV-cache region size ($n = n_{\text{text}} + k$)
    for High$\rightarrow$Low pitch conversion.}
    \label{fig:ablation_heatmap_high2low}
\end{figure}

Figure~\ref{fig:ablation_heatmap_high2low} visualizes the pitch difference (High$\rightarrow$Low) for each configuration. 
Negative values indicate successful pitch reduction toward the target style.
The results validate two key design decisions in our method.
When swapping only the text region ($k = 0$), all window sizes produce reversed pitch changes ($+6.8$ to $+16.2$ Hz), indicating that the KV-cache corresponding to the text region alone lacks sufficient target style information.
However, extending the swap to include the initial KV-cache ($k \geq 32$) for acoustic tokens enables successful conversion across all window sizes ($-5.3$ to $-12.4$ Hz).
This supports our findings that the model encodes critical style information in the initial KV-cache positions beyond the prompt, and our KV-cache swap must include these positions (i.e., $k > 0$) to achieve effective transitions.

\subsection{Adaptive Window Schedule}
In a separate test, we use $w=256$ for the first 3 seconds after $t^*$ and then enlarge the window. For Low-to-High pitch conversion, keeping $w=256$ throughout yields a $+30.5$ Hz change and SIM of 0.80. Enlarging the window to 512 after 3 seconds yields $+22.7$ Hz and SIM of 0.86, while switching to full attention yields $+14.0$ Hz and SIM of 0.90. For Quick-to-Slow, the adaptive schedule retains the measured speed change while increasing SIM from 0.81 to 0.88. These results illustrate a trade-off between transition strength and speaker similarity.

\begin{figure*}[t]
    \centering
    \includegraphics[width=\textwidth]{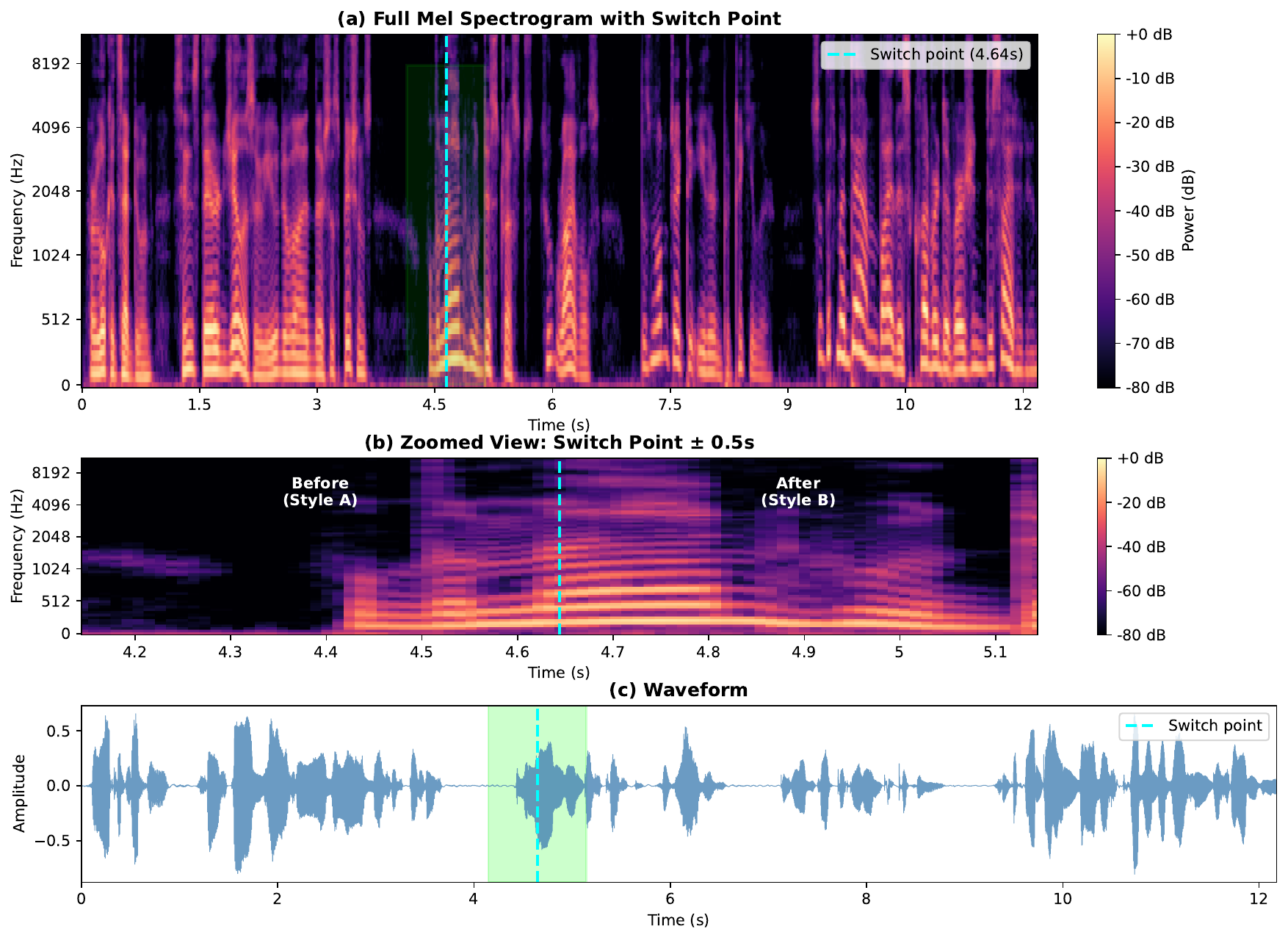}
    \caption{Spectrogram and waveform analysis of a generated sample with intra-utterance style transition. (a) Full mel-spectrogram with the switch point (cyan dashed line). (b) Zoomed-in view of the switch point $\pm 0.5$s. (c) Corresponding waveform; the green-shaded region marks the transition area. No vertical click line or amplitude discontinuity is observed at the boundary.}
    \label{fig:spectrogram_transition}
\end{figure*}

\begin{algorithm*}[t!]
\caption{Intra-utterance Style Transition via Direction Vector and KV-cache Swap}
\label{alg:intra_style}
\begin{algorithmic}[1]
\REQUIRE Text prompt $P$, source style text $S$, target style text $T$, source keyword, target keyword, transition point $t^*$, window size $w$, prefix steps $k$, strength $\alpha$
\ENSURE Generated audio with style transition

\STATE \textit{// Extract direction vector}
\STATE Extract direction vector $\mathbf{d}$ from $S$ and $T$ at keyword positions
\STATE $n \leftarrow |P| + k$ \hfill // text length + prefix steps

\STATE \textit{// Initialize decoders with style embeddings}
\STATE Initialize Decoder A with source style embeddings $\mathbf{E}^{(s)}$
\STATE Apply direction vector: $\mathbf{E}^{'} \leftarrow \mathbf{E}^{(s)} + \alpha \cdot \mathbf{d}$ at keyword position
\STATE Initialize Decoder B with target style embeddings $\mathbf{E}^{'}$

\STATE \textit{// Phase 1: Pre-compute target KV-cache}
\STATE Prefill Decoder B with text prompt $P$
\FOR{$t = 1$ \TO $k$}
    \STATE Generate an audio token with Decoder B
\ENDFOR
\STATE Store the resulting $\mathbf{K}^{(B)}_{1:n}, \mathbf{V}^{(B)}_{1:n}$

\STATE \textit{// Phase 2: Generate until transition with standard causal attention}
\FOR{$t = 1$ \TO $t^*$}
    \STATE Generate audio token with Decoder A using standard causal self-attention
\ENDFOR

\STATE \textit{// Phase 3: Swap at transition point}
\STATE $\mathbf{K}^{(A)}_{1:n}, \mathbf{V}^{(A)}_{1:n} \leftarrow \mathbf{K}^{(B)}_{1:n}, \mathbf{V}^{(B)}_{1:n}$ \hfill $\triangleright$ Eq.~\ref{eq:kv_swap}
\STATE $\mathbf{E}^{(s)} \leftarrow \mathbf{E}^{'}$

\STATE \textit{// Phase 4: Continue generation with sliding window}
\FOR{$t = t^* + 1$ \TO end}
    \STATE Apply sliding window mask $M[i,j]$ with window size $w$ \hfill $\triangleright$ Eq.~\ref{eq:sliding_mask}
    \STATE Generate audio token with Decoder A
\ENDFOR

\RETURN Generated audio sequence
\end{algorithmic}
\end{algorithm*}

\section{Algorithm for Intra-utterance Style Variation}
\label{appendix:algorithm}

Algorithm~\ref{alg:intra_style} provides the complete procedure for our intra-utterance style variation method, combining direction vector interpolation, a dual-decoder setup, KV-cache swap, and sliding window attention masking.

\section{Spectrogram Analysis at Transition Point}
\label{appendix:spectrogram}

To verify that our intra-utterance style transition produces smooth audio without boundary artifacts (e.g., clicks or discontinuities) potentially introduced by sliding-window attention masking, we inspect the mel-spectrogram and waveform around the transition point. As shown in Figure~\ref{fig:spectrogram_transition}, the zoomed mel-spectrogram (b) exhibits no vertical streaks or sudden energy bursts at the switch point, and the waveform (c) shows a continuous amplitude envelope across the transition region. This indicates that KV-cache swapping combined with sliding-window masking enables stylistic change without perceptible boundary artifacts, consistent with the high smoothness scores ($\geq 3.48$) reported in Table~\ref{tab:intra_results}.

\end{document}